\documentclass[10pt,journal,compsoc]{IEEEtran}

\ifCLASSOPTIONcompsoc
  \usepackage[nocompress]{cite}
\else
  \usepackage{cite}
\fi

\ifCLASSINFOpdf
\else
\fi
\usepackage{wrapfig}
\usepackage{graphicx}
\usepackage[table]{xcolor}
\definecolor{LightCyan}{rgb}{0.88,1,1}
\definecolor{LightYellow}{rgb}{1,1,0.7}
\usepackage{verbatim}
\usepackage{amsmath}
\usepackage{hyperref}
\usepackage[percent]{overpic}
\usepackage{pifont}

\def\eg{\emph{e.g.}}

\def\ie{\emph{i.e.}}
\def\etal{\emph{et al.}}
\def\kitti{KITTI}
\def\dispnet{DispNetC}
\newcommand{\xmark}{\ding{55}}%

\begin{document}
%
\title{Unsupervised Domain Adaptation for \\ Depth Prediction from Images}

\author{Alessio~Tonioni*,~\IEEEmembership{Student~Member,~IEEE,}
        Matteo~Poggi*,~\IEEEmembership{Member,~IEEE,}\\
        Stefano~Mattoccia,~\IEEEmembership{Member,~IEEE,}
        and~Luigi~Di~Stefano,~\IEEEmembership{Member,~IEEE}
        \IEEEcompsocitemizethanks{\IEEEcompsocthanksitem *joint first authorship}
\IEEEcompsocitemizethanks{\IEEEcompsocthanksitem A. Tonioni, M. Poggi, S. Mattoccia and L. Di Stefano are with the Department
of Computer Science and Engineering, University of Bologna, Italy,
IT.\protect\\
$\lbrace$alessio.tonioni,m.poggi,stefano.mattoccia,luigi.distefano $\rbrace$ @unibo.it
}
}

\IEEEtitleabstractindextext{%
\begin{abstract}

State-of-the-art approaches to infer dense depth measurements from images rely on CNNs trained end-to-end on a vast amount of data. However, these approaches suffer a drastic drop in accuracy when dealing with environments much different in appearance and/or context from those observed at training time. This domain shift issue is usually addressed by fine-tuning on smaller sets of images from the target domain annotated with depth labels. Unfortunately, relying on such supervised labeling is seldom feasible in most practical settings. 
Therefore, we propose an unsupervised domain adaptation technique which does not require groundtruth labels. Our method relies only on image pairs and leverages on classical stereo algorithms to produce disparity measurements alongside with confidence estimators to assess upon their reliability.
We propose to fine-tune both \emph{depth-from-stereo} as well as \emph{depth-from-mono} architectures by a novel confidence-guided loss function that handles the measured disparities as noisy labels weighted according to the estimated confidence.  
Extensive experimental results based on standard datasets and evaluation protocols prove that our technique can address effectively the domain shift issue with both  stereo and monocular depth prediction architectures  and  outperforms  other state-of-the-art unsupervised loss functions that may be alternatively deployed to pursue domain adaptation.

\end{abstract}

}

\maketitle

\IEEEdisplaynontitleabstractindextext

%
\IEEEpeerreviewmaketitle

\IEEEraisesectionheading{\section{Introduction}\label{sec:introduction}}

Depth sensing plays a central role in many computer vision applications. Indeed, the availability of 3D data can boost the effectiveness of solutions to tasks as relevant as autonomous or assisted driving, SLAM, robot navigation and guidance, and many others.
Active 3D sensors exhibit well-known drawbacks that may limit their practical usability: LiDAR, \eg, is cumbersome, expensive and provides only sparse measurements, while structured light features a limited working range and is mainly suited to indoor environments. On the other hand, passive techniques enabling to infer depth from images are suitable to most scenarios due to their low cost and easiness of deployment.
Among these, binocular stereo \cite{scharstein2002taxonomy} represents one of the most popular choices and a very active research topic since several decades.
Depth-from-stereo relies on finding the displacement (disparity) between corresponding pixels in two horizontally-aligned frames, which, in turn, enables depth estimation via triangulation.  Although stereo has been tackled for years by hand-engineered algorithms, deep learning approaches have recently proved to be effective and yield superior accuracy. The advent of deep learning in stereo initially concerned replacing key steps within traditionally handcrafted pipelines. Afterward, the whole process was addressed by deep architectures trained end-to-end to regress depths (disparities) from image pairs. These approaches represent nowadays the undisputed state-of-the-art provided that a vast amount of stereo pairs endowed with groundtruth depth labels are available for training. Purposely, the training procedure for end-to-end stereo architectures relies on an initial optimization based on a large synthetic dataset \cite{Mayer_2016_CVPR} followed by fine-tuning on, possibly many, image pairs with groundtruth sourced from the target domain. As a matter of fact, the popular \kitti{}  benchmarks \cite{KITTI_2012,KITTI_2015} witness the supremacy of deep stereo architectures \cite{yin2018hierarchical,cheng2018learning}, while this is quite less evident in the Middlebury benchmark \cite{MIDDLEBURY_2014}, where traditional, hand-crafted algorithms \cite{Taniai18,li20173d} still keep the top rankings on the leaderboards due to the smaller amount of images available for training. 
Deep learning did also dramatically boost development and performance of depth-from-mono architectures, which can predict depth from just one image and, thus, be potentially deployed on the far broader range of devices equipped with a single camera. 

Nonetheless, with both stereo and monocular setups, deep architectures aimed at predicting depth from images are severely affected by the \emph{domain shift} issue, which hinders effectiveness when performing inference on images significantly diverse from those deployed throughout the training process.  This can be observed,  for instance, when moving between indoor and outdoor environments, from synthetic to real data or between different outdoor/indoor environments. As already pointed out, in the standard training procedure this issue is addressed by fine-tuning on labeled images from the target domain. However, suitable labeled data are available only for a few benchmark datasets, \eg{} KITTI, whilst in most practical settings acquiring images annotated by depth labels would require the deployment of expensive sensors (\eg,  LiDAR) alongside with careful calibration. As this procedure is cumbersome and costly,  collecting and labeling enough images to pursue fine-tuning in the target domain may easily turn out unfeasible.  
Thus, although all state-of-the-art approaches for depth/disparity estimation from images rely on deep CNNs, the domain shift issue prevents widespread adoption of these architectures in practical settings. 

To address the above issue, in this paper we propose an unsupervised technique which allows for fine-tuning end-to-end architectures aimed at depth prediction without the need for groundtruth labels from the target domain. 
We argue that classical stereo matching algorithms rely on domain-agnostic computations that can deliver disparity/depth measurements in any working environment seamlessly. Although these measurements are prone to errors due to the known sub-optimality of stereo algorithms, we posit that they may be deployed as noisy labels to pursue fine-tuning of depth prediction architectures. Indeed, state-of-the-art estimators can reliably assess the confidence of disparity/depth predictions. Thus, we propose a novel learning framework based on a confidence-guided loss function which allows for fine-tuning depth prediction models by weighting the disparity/depth measurements provided by a stereo algorithm according to the estimated confidence.
As a result, our approach can perform adaptation by solely feeding the model with synchronized stereo images from the target domain, \ie{} without requiring cumbersome and expensive depth annotations.

{\section{Related work}\label{sec:related}}

\textbf{Deep stereo.} Since the early works on stereo, classical algorithms \cite{scharstein2002taxonomy} comprise several sequential steps dealing with initial matching cost computation, local aggregation, disparity optimization and refinement.
The first attempt to plug deep learning into a well established stereo pipeline was aimed at replacing matching cost computation \cite{zbontar2015computing,Chen_2015_ICCV,luo2016efficient}, while disparity optimization \cite{Seki_2016_BMVC,Seki_2017_CVPR} and refinement \cite{Gidaris_2017_CVPR} have been addressed more recently. 
Although these works proved the superiority of learning-based methods in the addressed steps, in most cases traditional optimization strategies, such as Semi Global Matching (SGM) \cite{hirschmuller2005accurate}, were needed to reach top accuracy.
The shift toward end-to-end architectures started with DispNet, a seminal work by Mayer et al. \cite{Mayer_2016_CVPR}. Unlike previous proposals that process small image patches to compute similarity scores \cite{zbontar2015computing,Chen_2015_ICCV,luo2016efficient}, \dispnet{} relies on a much larger receptive field, extracts features jointly from the two input images and computes correlations to predict the final disparities. This approach, however,  mandates a significant amount of labeled training samples such that  the few hundreds of images available in  \kitti{} \cite{KITTI_2012,KITTI_2015} turn out definitely insufficient. To tackle this issue, a large synthetic dataset \cite{Mayer_2016_CVPR} was created and deployed  for training, with \kitti{} images  used to address the domain shift issue arising when running the network on real imagery. Although \dispnet{} did not reach the top rank on \kitti{}, it inspired other end-to-end models \cite{Pang_2017_ICCV_Workshops,Liang_2018_CVPR,yin2018hierarchical} which, in turn, were able to achieve state-of-the-art performance. Along a similar research line, some authors deploy 3D convolutions to exploit geometry and context   \cite{Kendall_2017_ICCV,AAAI1816467,Chang_2018_CVPR,cheng2018learning}.
Despite the different architectural details, these techniques follow the same synthetic-to-real training schedule as originally proposed for DispNet. Differently, Zhout et. al. \cite{zhou2017unsupervisedb} described an iterative procedure based on the left-right check to train a deep stereo network from scratch without the need of groundtruth disparity labels. Finally, Zhang et al. \cite{zhang2018activestereonet} proposed a novel loss function formulation to enable depth estimation without supervision within an active stereo acquisition setup.

\textbf{Confidence measures for stereo.} Confidence measures were extensively reviewed at first by Hu and Mordohai \cite{Hu_2012_PAMI} and more recently by Poggi et al. \cite{Poggi_2017_ICCV}, who considered approaches leveraging on machine-learning. These are mainly based either on random forests   \cite{Hausler_2013_CVPR,Spyropoulos_2014_CVPR,Park_2015_CVPR,Poggi_2016_3DV} or CNNs  \cite{Poggi_2016_BMVC,Seki_2016_BMVC,Fu_2018_WACV,TOSI_2018_ECCV}. While most of the former methods usually combine different cues available from the intermediate cost volume calculated by classical stereo algorithms \cite{Secaucus_1994_ECCV,hirschmuller2005accurate,zbontar2016stereo}, the latter can deploy just disparity maps and image cues, which renders it amenable also to depth estimation frameworks, such as end-to-end CNNs, that do not explicitly provide a cost volume. Moreover, CNN-based confidence estimators have been recently shown to exhibit better outlier detection performance  \cite{Poggi_2017_ICCV}.
\cite{Poggi_2017_CVPR}  proposed an effective deep learning approach to improve confidence measures by exploiting local consistency while \cite{Poggi_2017_CVPR_workshops} a method to ameliorate random forest-based approaches for confidence fusion \cite{Spyropoulos_2014_CVPR,Park_2015_CVPR,Poggi_2016_3DV}. Shaked and Wolf \cite{Shaked_2017_CVPR} embedded confidence estimation within a deep stereo network while other works looked deeper into the learning process of confidence measures, either by studying features augmentation \cite{kim2017feature} or designing self-supervised techniques to train on static video sequences \cite{MOSTEGEL_CVPR_2016} or stereo pairs \cite{Tosi_2017_BMVC}. 
Finally, Poggi et al. \cite{Poggi_2017_ICIAP} evaluated simplified confidence measures for embedded systems.

\textbf{Depth-from-mono.} Deep learning dramatically boosted the results attainable by a monocular depth prediction setup. While the vast majority of works addressed the depth-from-mono problem through supervised learning \cite{saxena2009make3d,ladicky2014pulling,eigen2014depth,liu2016learning,laina2016deeper,li2015depth,ummenhofer2017demon,fu2018supervised, xu2018supervised}, an exciting recent trend concerns self-supervising the model by casting training as an image reconstruction problem. 
This formulation is earning increasing attention due to the potential to train depth prediction networks without hard to source depth labels. 
Self-supervised depth-from-mono methods can be broadly classified into monocular and stereo. With the former approach \cite{zhou2017unsupervised,mahjourian2018unsupervised,wang2018unsupervised,yin2018geonet} images are acquired by an unconstrained moving camera and the estimated depth is used to reconstruct views across the different frames through camera-to-world projection and vice-versa. Thus, the network has to estimate also the unknown camera pose between frames and the computation tends to fail when moving objects are present in the scene. 
The latter  category requires a calibrated stereo setup to carry out the training phase \cite{garg2016unsupervised,godard2017unsupervised,pydnet18,zhan2018unsupervised, Aleotti_monogan_2018}. As, in this case, the relative pose between the two cameras is known, the network has only to estimate the depth (actually, disparity) that minimizes the reprojection error between the two views. Thus, on one hand, this strategy can handle seamlessly moving objects, on the other it constraints data collection. Networks trained according to a stereo setup yield usually more accurate depth estimations. Moreover, this approach can be extended to three views  \cite{3net18} to compensate for the occlusions inherited by the binocular setup. 
Finally, we mention the joint use of these two supervision strategy \cite{zhan2018unsupervised} and the semi-supervised frameworks proposed in \cite{Kuznietsov_2017_CVPR,kumar2018gan} that combined sparse groundtruth labels with stereo supervision.

In \cite{Tonioni_2017_ICCV} we highlighted the issues and challenges set forth by the deployment of deep stereo architectures across multiple domains due to the lack of labeled data to perform fine-tuning. Accordingly, we proposed to adapt a deep stereo network to a new domain without any supervision by a novel loss function that leverages on a confidence estimator in order to detect reliable measurements among the disparities provided by a classical stereo algorithm.  Later, Pang et al. \cite{Pang_2018_CVPR} addressed the same topic and proposed to achieve adaptation of a deep stereo network by combining the disparity maps computed at multiple resolutions within an iterative optimization procedure.  

This paper extends the early ideas and findings presented in  \cite{Tonioni_2017_ICCV}. In particular, while in \cite{Tonioni_2017_ICCV} we considered only deep stereo,  we provide here a general formulation to addresses both depth-from-stereo as well as depth-from-mono. Besides, we present a more comprehensive collection of quantitative and comparative experimental results. As for depth-from-stereo, thanks to the vast amount of depth labels released recently  \cite{Uhrig2017THREEDV}, starting with \dispnet{} \cite{Mayer_2016_CVPR} pre-trained on synthetic data we show adaptation results on the \kitti{} raw dataset \cite{KITTI_RAW}, which includes more than 40k images. As for depth-from-mono, we  consider the deep architecture recently proposed by Godard et al. \cite{godard2017unsupervised} and perform  domain adaptation from the CityScapes dataset \cite{cordts2016cityscapes} toward  \kitti{}.

{\section{Domain adaptation for depth sensing}\label{sec:method}}

This section describes our domain adaptation framework, which is suited to both deep stereo as well as monocular depth estimation networks. To adapt a pre-trained model facing a new environment, we first acquire stereo pairs from the target domain. Then, we deploy a classical  (\ie, not learning-based) stereo algorithm to generate dense depth measurements together with a state-of-the-art confidence measure to estimate the reliability of the depth values calculated by the stereo algorithm. 

A key observation behind our method is that classical stereo algorithms, although affected by well-known shortcomings such as occlusions, poorly-textured regions, and repetitive patterns, are substantially agnostic to the specific target environment and thus behave similarly across different scenarios. More importantly, they fail in the same predictable way, thereby enabling confidence measures to achieve remarkably good accuracy in detecting mistakes regardless of the sensed environment \cite{Poggi_2017_ICCV}. 

Based on the above observations, we advocate deploying the depths delivered by a classical stereo algorithm as noisy labels endowed with reliability estimations in order to fine-tune a network aimed at depth prediction. This is achieved through a novel per-pixel regression loss wherein the error between each model prediction and the corresponding depth measurement provided by the stereo algorithm is weighted according to the reliability estimated by the confidence measure, with higher weights associated to more reliable depth measurements. Thereby, the learning process is guided by the high-confidence depth measurements, \ie{} those labels that appear to be more reliable, while the errors due to the shortcomings of the stereo algorithm have a negligible impact. 

Thus, given a pre-trained depth estimation network, either stereo or monocular, and a set of stereo pairs, $(I^l,I^r) \in \mathcal{I}$, acquired from the target domain, for each pair we compute a dense disparity map,  $D \in \mathcal{D}$, by means of a classical stereo algorithm, $f: (\mathcal{I},\mathcal{I})\rightarrow\mathcal{D}$, such as, \eg{},  SGM \cite{hirschmuller2005accurate} or AD-CENSUS\cite{Secaucus_1994_ECCV}. Moreover, for each disparity map, $D$, we estimate a pixel-wise degree of reliability according to a confidence measure, $c: \mathcal{D} \rightarrow \mathcal{C}$. The resulting confidence map, $C \in \mathcal{C}$, encodes the reliability of the disparity calculated at each pixel as a score ranging from $0$ (\emph{not reliable}) to $1$ (\emph{reliable}).

We run $f$ and $c$ on each stereo pair available from the target domain so as to produce the training set deployed to perform fine-tuning of the pre-trained depth estimation network. Therefore, each sample, ($S_i$), in the training set is a tuple of four elements:

\begin{equation}
    \label{eq:training set}
    S_i=(I^l_i,I^r_i,D_i,C_i)=(I^l_i,I^r_i,f(I^l_i,I^r_i),c(f(I^l_i,I^r_i))) 
\end{equation}

Given the depth estimation network (either stereo or monocular), which takes input images and outputs per pixel disparities, we fine tune it toward the target domain by minimizing a loss function, $L$, consisting of three terms:  a \textit{confidence guided loss} ($L_c$), a \textit{smoothing loss} ($L_s$) and an \textit{image reconstruction loss} ($L_r$):

\begin{equation}
    \label{eq:loss}
    L =  L_c + \lambda_1 \cdot L_s + \lambda_2 \cdot L_r
\end{equation}
with $\lambda_1,\lambda_2$  hyper-parameters to weight the contribution of the associated loss terms. All the three components of our loss can be applied seamlessly to deep learning models aimed either at depth-from-stereo or depth-from-mono (in the latter case one just need to convert disparities into depths). 
The structure of the three terms in \autoref{eq:loss} is detailed in the next sub-sections, while
in \autoref{sec:experiments} we present model ablation experiments aimed at assessing their individual contribution to performance. 

\subsection{Confidence Guided Loss}
\label{ssec:confidence_loss}

\begin{figure*}
\centering
\setlength{\tabcolsep}{1pt}
\begin{tabular}{cccc}
    \begin{overpic}[width=0.24\textwidth]{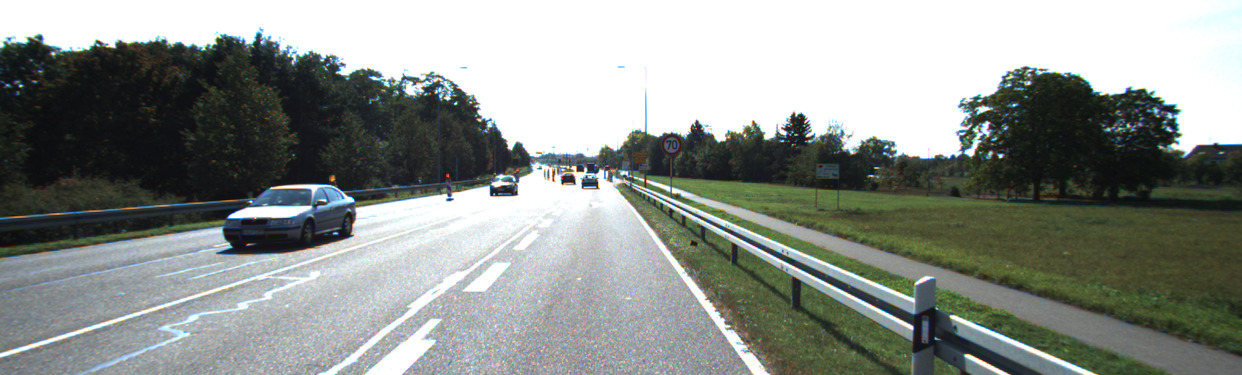}
    \end{overpic} &
    
    \begin{overpic}[width=0.24\textwidth]{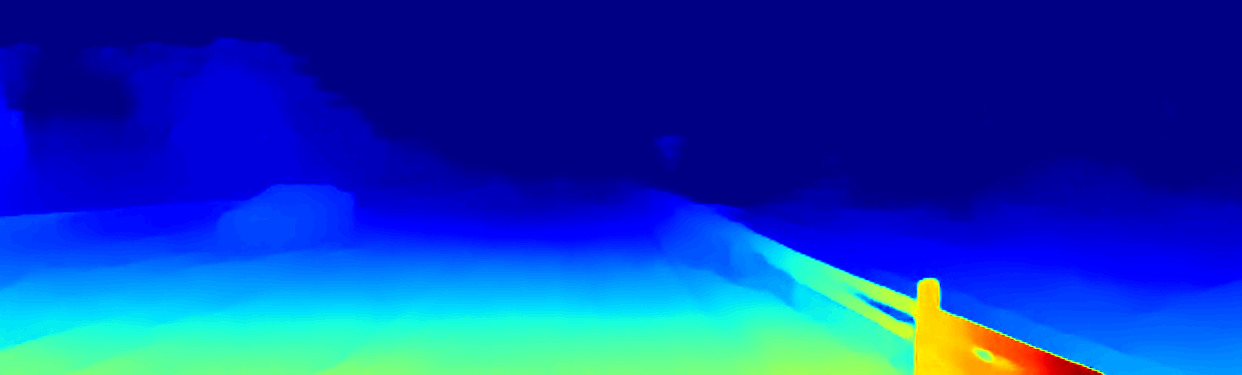}
    \end{overpic}&
    \begin{overpic}[width=0.24\textwidth]{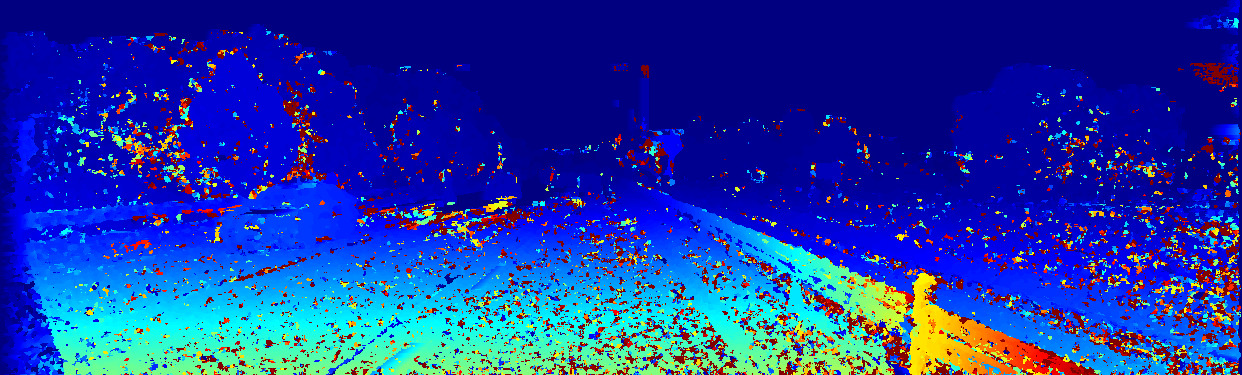}
    \end{overpic}&
    \begin{overpic}[width=0.24\textwidth]{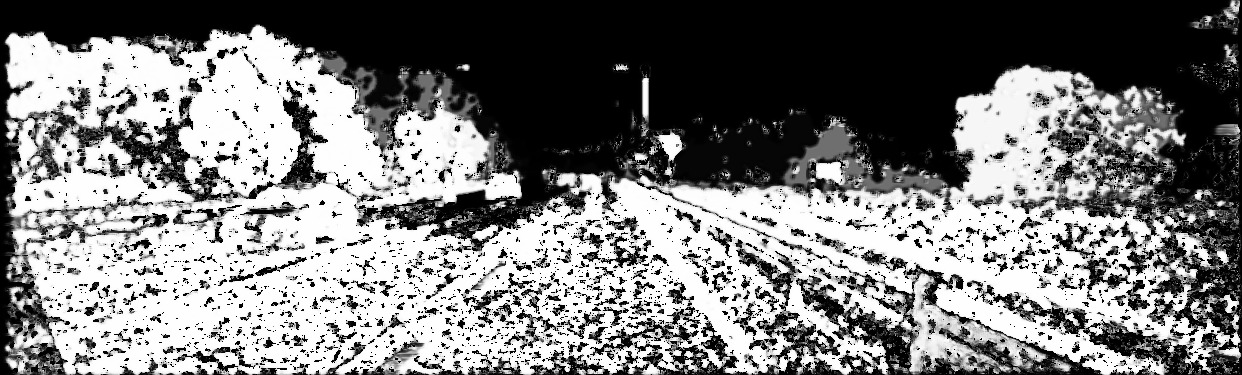}
    \end{overpic}
    \\
    (a) & (b) & (c) & (d) \\
    
    \begin{overpic}[width=0.24\textwidth]{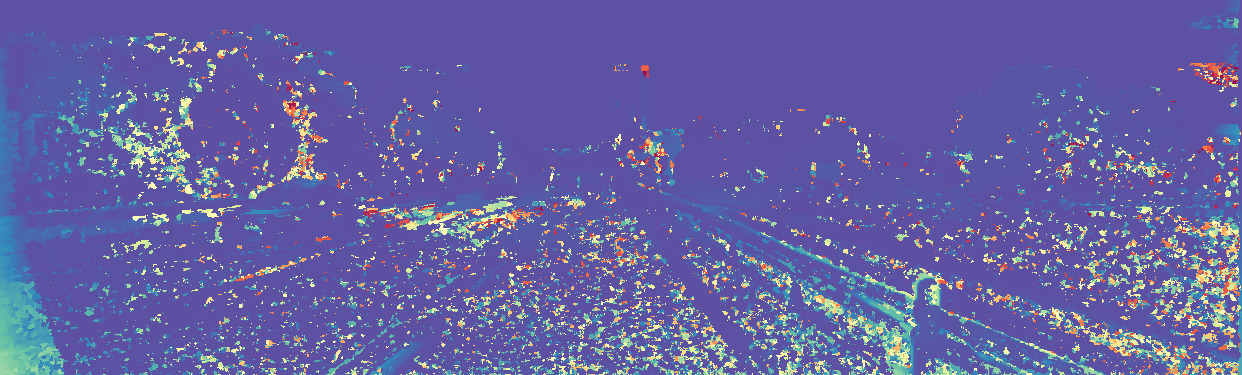}
    \end{overpic}&
    \begin{overpic}[width=0.24\textwidth]{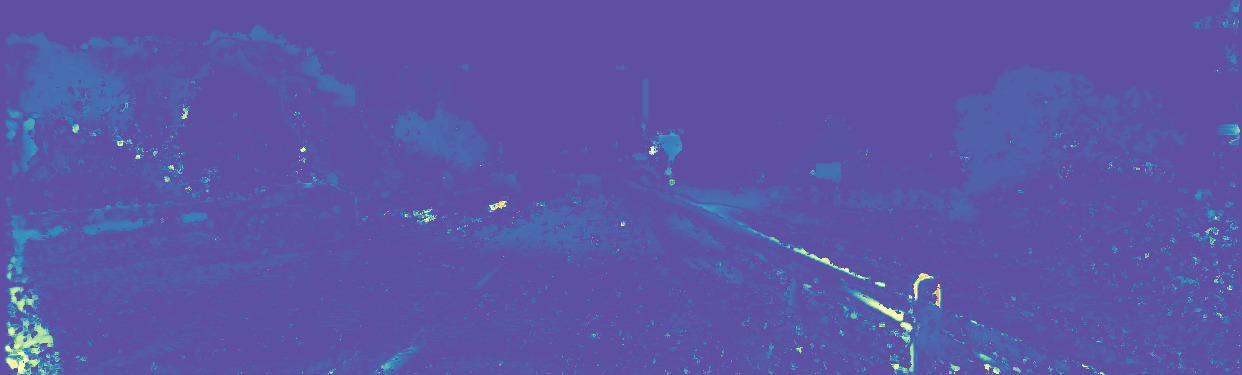}
    \end{overpic}&
      \begin{overpic}[width=0.24\textwidth]{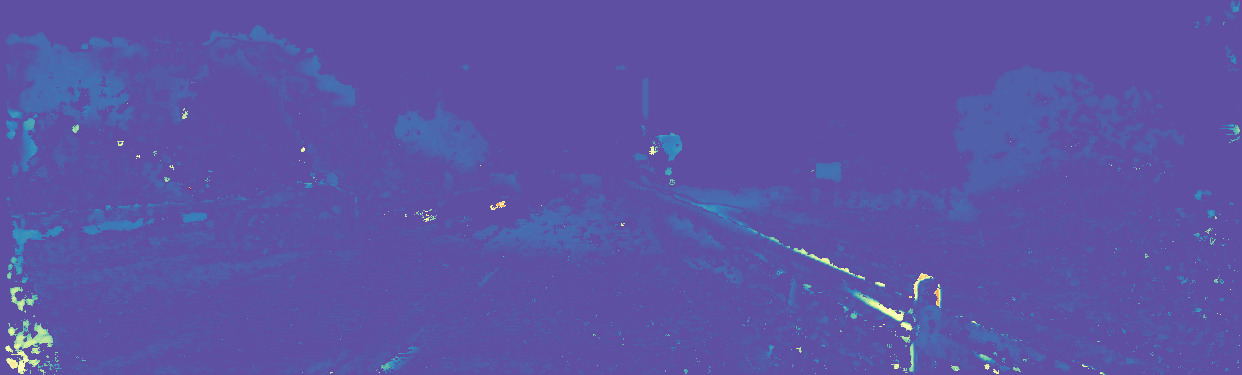}
    \end{overpic}&
    \begin{overpic}[width=0.24\textwidth]{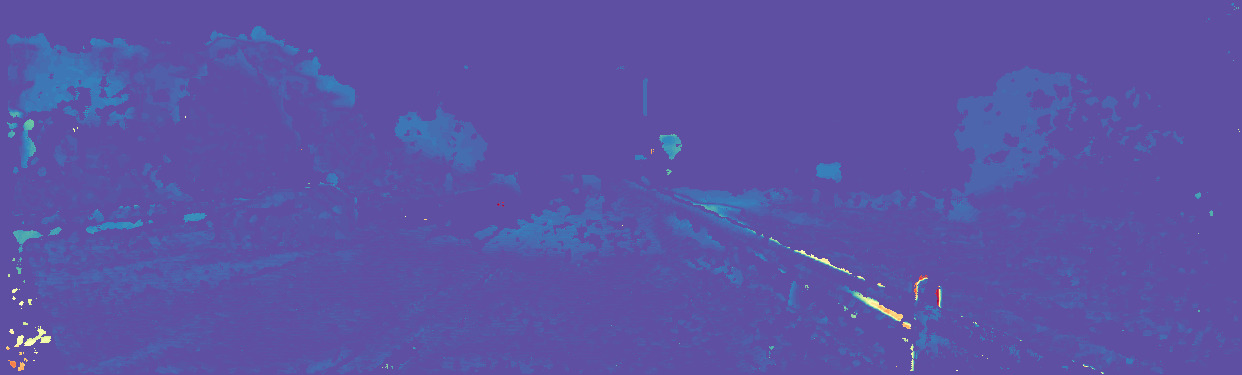} 
    \end{overpic}
    \\
    (e)  & (f)  & (g)  & (h)  \\
    
\end{tabular}
\caption{ Visualization of our confidence guided loss: (a) left frame $I^l$; (b) Disparity map, $\tilde{D}$, predicted by the model; (c) Disparity map, $D$, estimated by a stereo algorithm; (d) Confidence map, $C$, on $D$; (e) L1 regression errors between (b) and (c), (f-h) same L1 errors weighted by $C$ with $\tau=0.00$ (f), $\tau=0.50$ (g) and $\tau=0.99$ (h). (e-h) Hotter colors encode larger differences. 
}
\label{fig:tau_overlap}
\end{figure*}

The inspiration for the $L_c$ term in the loss function of \autoref{eq:loss} comes from the observation that deep models can be successfully fine-tuned to new environments even by deploying only a few sparse groundtruth annotations. This is vouched by the performance achievable on the \kitti{} datasets \cite{KITTI_2012,KITTI_2015,KITTI_RAW}, where only a subset of pixels carries depth annotations (roughly $\frac{1}{3}$ of the image). The common strategy to account for the missing values consists simply in setting the loss function to $0$ at those locations, thereby providing the network with meaningful gradients only at a subset of the spatial locations. Indeed,  even in these sub-optimal settings, networks are able to adapt and ameliorate accuracy remarkably well. 
We build on these observations and leverage on the confidence measure, $c$, to obtain sparse and reliable depth labels from the noisy output $D$ of the stereo algorithm. With reference to  \autoref{eq:training set}, denoting as $\tilde{D}$ the output predicted by the model at the current training iteration, we compute $L_c$ as

\begin{equation}
    \label{eq:confidence_loss}
    L_c = \frac{1}{|P_v|}\sum_{p \in \mathcal{P}_v} \mathcal{E}(p)
\end{equation}
\begin{equation}
    \label{eq:pixelwise_loss}
    \mathcal{E}(p) = C(p) \cdot |\tilde{D}(p)-D(p)|
\end{equation}
\begin{equation}
    \label{eq:valid_points}
    \mathcal{P}_v=\{p \in \mathcal{P}:C(p)>\tau\}
\end{equation}

where $\mathcal{P}$ is the set of all spatial locations on the image and $\tau \in [0,1]$ a hyper-parameter that controls the sparseness and reliability of the disparity measurements provided by $f$ that are deployed to update the model. A higher value of $\tau$ will mask out more mistakes in $D$ though permitting less spatial locations to contribute to model update. Hence, points belonging to $\mathcal{P}_v$  define a set of sparse labels that, assuming the availability of a perfect confidence measure, may be used as if they were  \emph{groundtruth annotations}, \eg{} akin to the LiDAR measurements deployed in the \kitti{} dataset. Yet, confidence measures are not perfect and often show some degree of uncertainty in the score assigned to disparity measurements. Thus, we weight the contribution at location $p$ by $C(p) \in [0,1]$, \ie{} as much as the depth measurement, $D(p)$, can be trusted according to the confidence estimation, $C(p)$. We point out that, re-weighting the loss function in the presence of noisy labels has been successfully exploited in supervised classification \cite{ren2018learning,jiang2017mentornet}. Our formulation deploys a similar idea for a dense regression problem. Yet, we leverage on an external and highly accurate strategy to detect noise in the labels (\ie{}, the confidence measure) and mask out those labels which, according to the adopted strategy, are very likely wrong, \ie{}, $\{D(p): p \notin \mathcal{P}_v\}$. In \autoref{sssec:ablation} we will show how both masking and re-weighting are crucial components to maximize performance in the presence of noisy depth labels. 

The bottom row of \autoref{fig:tau_overlap}  shows a graphical visualization of the errors that our $L_c$ loss term tries to minimize. On (e) we report the errors that will be minimized trying to directly regress the noisy depth labels of (c) given the model prediction on (b); on (f-g-h), instead, the errors minimized by applying $L_c$ with different $\tau$ values ($0,0.5$ and $0.99$ respectively). By tuning $\tau$ we can control the number of pixels, and therefore labels, taking part in the network adaptation process. Clearly, leveraging on  more  labels comes at the cost of injecting more noise in the process, which, in turn, may harm  adaptation, even if their contribution will be attenuated by $C$, \eg{} compare (f) to (e) where the only difference is the scaling of errors by $C(p)$ in (f). In (h) we can appreciate how even with $\tau=0.99$ the amount of pixels considered during the optimization process is still quite high.  
We refer the reader to \cite{Tonioni_2017_ICCV} for a detailed analysis of the quantity and quality of the labels used in the optimization process for different values of $\tau$. 

\subsection{Self-filtering Outliers}
\label{ssec:learning_tau}

In our previous work, \cite{Tonioni_2017_ICCV}, a properly hand-tuned $\tau$ proved to be effective. However, as $\tau$  represents a hyper-parameter of the method, an appealing alternative would consist in learning it alongside with the model adaptation process. 
To this aim, we define $\tau$ as a learnable parameter in our framework and update its value by gradient descent anytime the confidence guided loss described in \autoref{ssec:confidence_loss} is optimized. 
Unfortunately, as $\tau$ determines the number of pixels on which such loss is computed, with this learning strategy its value would rapidly converge to 1, \ie{} so as to mask out all pixels in order to obtain a loss as small as zero. To avoid such a behavior, we reformulate \autoref{eq:confidence_loss} as

\begin{equation}
    \label{eq:tau_loss}
    L_c = \frac{1}{|P_v|}\sum_{p \in \mathcal{P}_v} \mathcal{E}(p) - \log{(1 - \tau)}
\end{equation}

The additional logarithmic penalty discourages $\tau$ from being equal to 1, thereby avoiding complete masking out of all pixels. 
In the experimental results, we will show how learning $\tau$ performs almost equivalently to the use of a hand-tuned threshold obtained by validation on ground-truth data. The latter, however, would turn out quite a less practical approach in those scenarios for which our adaptation technique is designed.
In our evaluation, we will report two main experiments by formulating $\tau$ as i) a learnable variable or ii) the output of a shallow neural network, referred to as $\tau$Net, applied to the reference image and consisting of  three $3\times3$ Conv layers with 64 filters followed by a global average pooling operation. With this second approach, we allow $\tau$ to be a function of the current image content rather than a fixed threshold for the whole dataset.

\subsection{Smoothing Loss}
\label{ss:smoothing_temr}

As $L_c$ produces error signals to improve disparity prediction only at the subset of sparse image locations $P_v$,  similarly to \cite{heise2013pm} we use an additional loss term, $L_s$, to propagate model update signals across neighboring spatial locations. In particular, $L_s$ tends to penalize large gradients  in the predicted disparity map ($\partial \tilde{D}$) while taking into account the presence of gradients in pixel intensities ($\partial I$):

\begin{equation}
    \label{eq:smoothness}
    L_s = \frac{1}{|P|} \sum_{p \in \mathcal{P}} \partial_x \tilde{D}(p)\cdot e^{-||\partial_x I(p)||} + \partial_y \tilde{D}(p)\cdot e^{-||\partial_y I(p)||}
\end{equation}

Thus, based on the consideration that depth discontinuities are likely to occur in correspondence of image edges, $L_s$ constrains the predicted disparity map, $\tilde{D}$, to be smooth everywhere but at image edges. To efficiently compute gradients along $x$ and $y$ we use convolutions with  $3 \times 3$ Sobel filter. 

\subsection{Image Reconstruction Loss}
\label{ss:image_reconstruction}

To further compensate for the sparse model update information yielded by $L_c$, we include in the loss function a pixel-wise \emph{image reconstruction} term, denoted as  $L_r$ in \autoref{eq:loss}. Inclusion of this term in our loss has been inspired by \cite{godard2017unsupervised}, which has shown how deploying image re-projection between stereo frames can deliver a form of self-supervision to train a depth-from-mono network. Hence, given a stereo pair,  $I^l$ can be reconstructed from $I^r$ according to the current disparity prediction $\tilde{D}$  by employing a bilinear sampler in order to render the process locally differentiable. Denoted as $\tilde{I}^l$  the re-projection of $I^r$ according to $\tilde{D}$, we define the image reconstruction loss, $L_r$, as a weighted combination of the $L1$ norm and the single scale SSIM \cite{wang2004image}:
\begin{equation}
\label{eq:ssim}
    L_r = \frac{1}{|P|}\sum_{p \in P} \alpha \frac{1-SSIM(I^l(p),\tilde{I}^l(p))}{2} +(1-\alpha)|I^l(p)-\tilde{I}^l(p)|
\end{equation}
Similarly to  \cite{godard2017unsupervised}, we use a simplified SSIM based on a $3 \times 3$ block filter and set $\alpha=0.85$ throughout all our experiments. 

\section{Experimental results}
\label{sec:experiments}

In this section, we present a large corpus of experiments aimed at assessing the effectiveness of our proposed unsupervised domain adaptation framework. 
As already mentioned, although in the initial proposal \cite{Tonioni_2017_ICCV} our approach was concerned with deep stereo models only, in this paper we present a general formulation to adapt any architecture trained to predict dense depth maps provided that stereo pairs are available at training time. Therefore, we address two main settings:  i) adaptation of a deep stereo network and ii) adaptation of a depth-from-mono network.
As for the former, we carry out extensive experiments according to the protocol proposed in our previous work \cite{Tonioni_2017_ICCV}. At that time, experiments were limited to \kitti{} 2012 and 2015, whilst in this paper, we can consider the whole \kitti{} raw dataset \cite{KITTI_RAW}, which includes about 40K images, thanks to the groundtruth labels released recently in the official website \cite{Uhrig2017THREEDV}.
As for the latter evaluation scenario, we follow the standard protocol from the literature of self-supervised monocular depth estimation \cite{godard2017unsupervised}, which consists in splitting the \kitti{} raw data into train and test, as proposed by  Eigen et al. \cite{eigen2014depth}.

To deploy the  confidence guided loss described in \autoref{ssec:confidence_loss}, in our evaluation we consider two classical stereo algorithms: AD-CENSUS (shortened AD) \cite{Secaucus_1994_ECCV} and Semi-Global Matching (shortened \emph{SGM}) \cite{hirschmuller2005accurate} and leverage the implementations of \cite{spangenberg2014large}. 
We have selected these two popular algorithms because they show quite different behaviors. While AD tends to generate prediction errors in the form of small spikes in the disparity maps, the errors generated by SGM can often cause over-smoothing. Effectiveness with both types of error patterns may help testify the general validity of our proposal. Besides, while SGM may turn out remarkably accurate, AD is notoriously significantly more prone to errors, which, in our framework, leads to fewer disparity measurements used at training time to compute $L_c$ due to fewer pixels belonging to $\mathcal{P}_v$. To measure the confidence of the disparity measurements coming from the stereo algorithms, we rely on CCNN \cite{Poggi_2016_BMVC} as it can yield state-of-the-art performance and does require just the disparity map as input. Thanks to the latter trait, CCNN can be applied to any stereo system, even in case one has no access to the source code of the algorithm or is willing to employ an off-the-shelf external device. As CCNN consists of a network trained to classify each disparity pixel as reliable or not according to a small support region, it needs to be trained before deployment. To avoid reliance on expensive depth annotations, we used the original authors' implementation\footnote{https://github.com/fabiotosi92/CCNN-Tensorflow} and trained two variants of the network - one for AD and the other for SGM - on synthetic images taken from the SceneFlow dataset \cite{Mayer_2016_CVPR}. More precisely,  we took six random stereo pairs from the Driving portion of the dataset (0040, 0265 forward from 15mm focal length set and 0075 forward, 0099, 0122, 0260 backward from 35mm set) and trained CCNN for 14 epochs, as suggested in \cite{Poggi_2016_BMVC}.

All the code developed is available to ease development of applications relying on depth sensing using deep learning models.\footnote{\url{https://github.com/CVLAB-Unibo/Unsupervised_Depth_Adaptation}} 

\begin{table*}[t]
\centering
\begin{tabular}{|l|ccc|cc|cc|}
\cline{2-8}
\multicolumn{1}{c|}{} & \multicolumn{3}{c|}{Hyper parameters} & \multicolumn{2}{c|}{Target Domain} & \multicolumn{2}{c|}{Similar Domains} \\
\hline
Test & $\tau$ & $\lambda_1$ & $\lambda_2$ & bad3 & MAE & bad3 & MAE \\
\hline
(a) AD \cite{Secaucus_1994_ECCV} & \xmark & \xmark & \xmark & 32.03 & 19.60 & 32.03 & 19.60 \\
(b) No Adaptation & \xmark & \xmark & \xmark & 10.86 & 1.73 & 10.86 & 1.73 \\
\hline
(c) Regression & \xmark & \xmark & \xmark & 11.73 & 2.49 & 12.23 & 2.47 \\
(d) Weighted & 0 & 0 & 0 & 3.66 & 1.03 & 4.57 & 1.12 \\
(e) Masked & 0.8 & 0 & 0 & 3.17 & 1.02 & 3.97 & 1.09 \\
(f) Masked+Smoothness & 0.8 & 0.1 & 0 & 3.17 & 0.98 & 3.78 & 1.05\\
(g) Masked+Reprojection & 0.8 & 0 & 0.1 & 3.03 & 0.98 & 3.70 & 1.05\\
(h) Complete Adaptation & 0.8 & 0.1 & 0.1 & \textbf{2.96} & \textbf{0.96} & \textbf{3.66} & \textbf{1.04} \\
(i) Learned Adaptation & learned & 0.1 & 0.1 & 3.15 & 1.01 & 3.84 & 1.08 \\
(j) \textit{$\tau$Net Adaptation} & learned & 0.1 & 0.1 & 3.15 & 0.99 & 3.83 & 1.07 \\
\hline
\end{tabular}
\caption{Ablation study on the effectiveness of the different components of our \emph{Adaptation} loss using AD as noisy labels estimator. Results computed on the \kitti{} RAW dataset using a 4-fold cross validation schema, best results highlighted in bold.}
\label{table:ablation}
\end{table*}

\begin{figure*}
\begin{tabular}{ccc}
\includegraphics[width=0.30\textwidth]{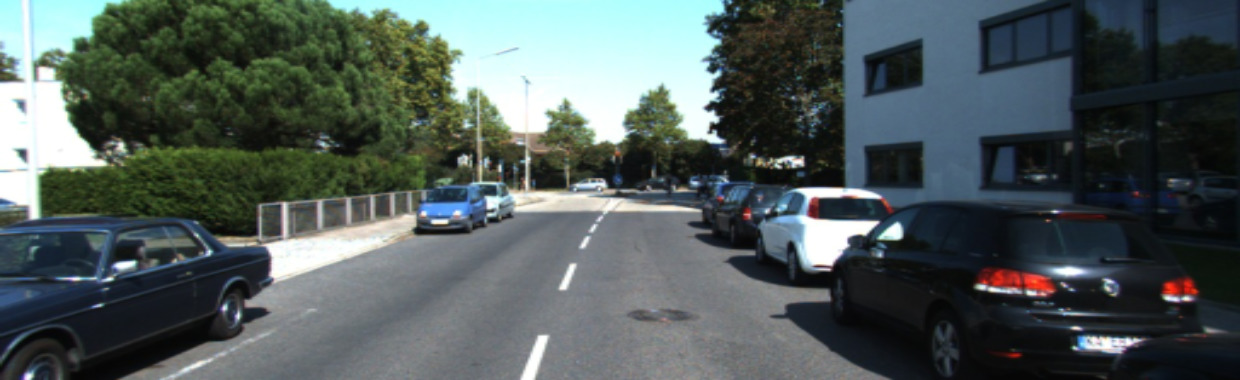} &
\begin{overpic}[width=0.30\textwidth]{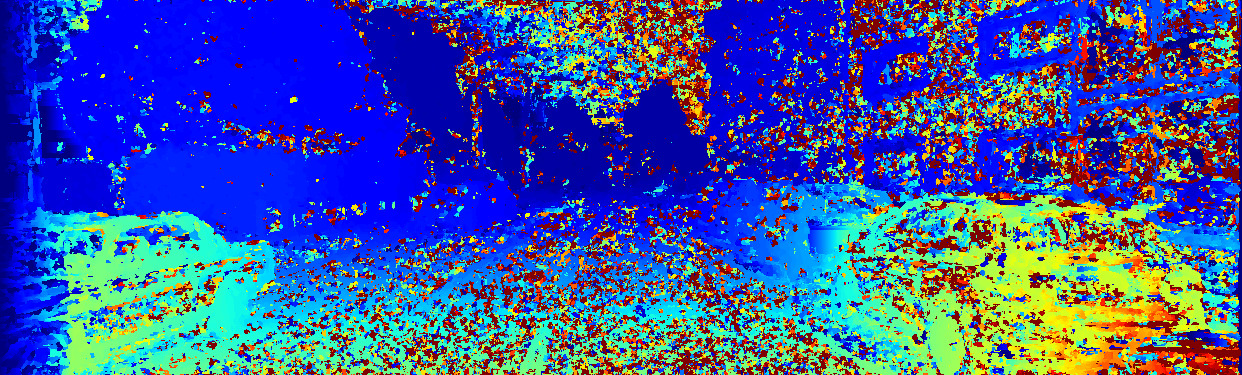}
\put (2,25) {$\displaystyle\textcolor{white}{\textbf{bad3: 38.12}}$}
\end{overpic} &
\begin{overpic}[width=0.30\textwidth]{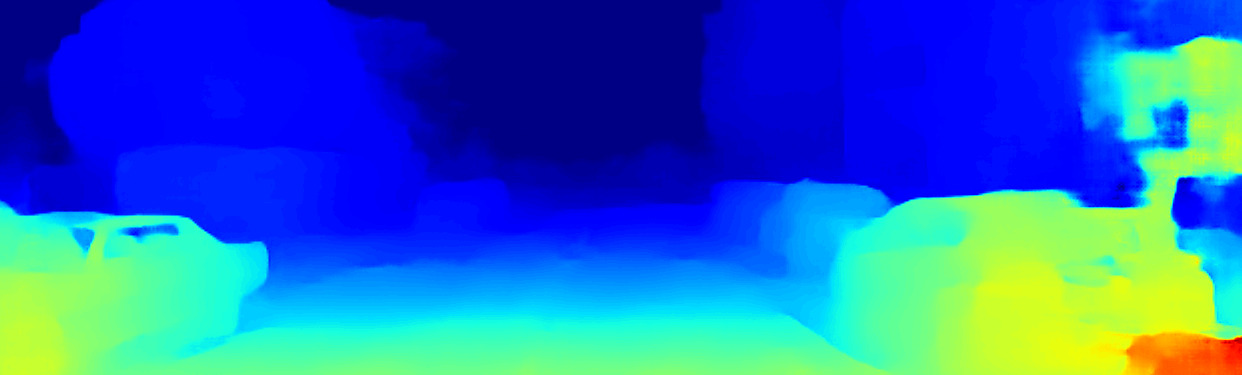}
\put (2,25) {$\displaystyle\textcolor{white}{\textbf{bad3: 3.56}}$}
\end{overpic} 
\\
(a) &
(b) &
(c) \\
\begin{overpic}[width=0.30\textwidth]{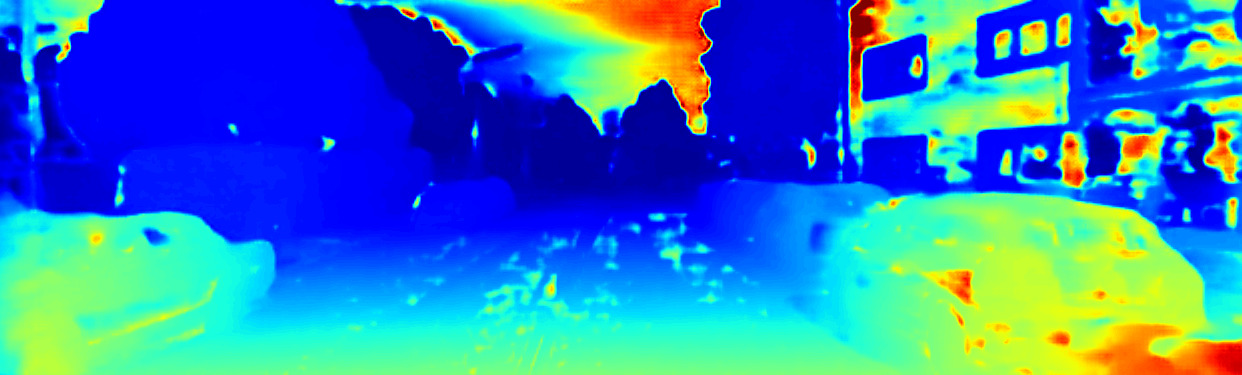}
\put (2,25) {$\displaystyle\textcolor{white}{\textbf{bad3: 12.06}}$}
\end{overpic} &
\begin{overpic}[width=0.30\textwidth]{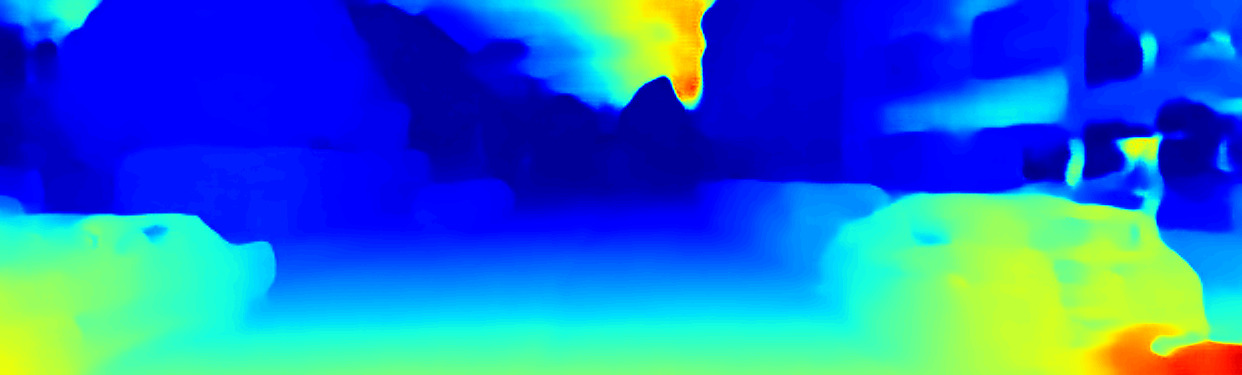}
\put (2,25) {$\displaystyle\textcolor{white}{\textbf{bad3: 1.54}}$}
\end{overpic} &
\begin{overpic}[width=0.30\textwidth]{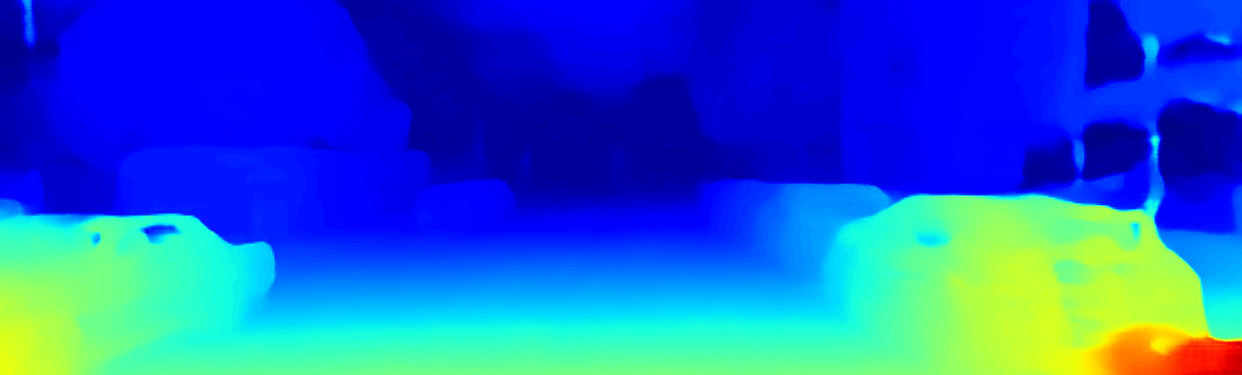}
\put (2,25) {$\displaystyle\textcolor{white}{\textbf{bad3: 1.35}}$}
\end{overpic}
\\
(d) & (e) & (f) \\
\end{tabular}
\caption{Ablation experiments: fine-tuning \dispnet{} to new domains using AD \cite{Secaucus_1994_ECCV}. (a) input image from \kitti{}, (b) disparities estimated by AD, (c) results without fine-tuning, (d) fine-tuning by AD only (\textit{Regression}), (e) fine-tuning by weighting the loss through the confidence estimator (\textit{Weighted}) and (f)  our complete \emph{Adaptation} method.}
\label{fig:stereo_ablation}
\end{figure*}

\subsection{Deep Stereo}
\label{ssec:stereo}
Our first experimental scenario is about the adaptation of a depth-from-stereo network to a new environment. The common training procedure for this kind of models consists of first training on the large synthetic FlyingThings3D dataset \cite{Mayer_2016_CVPR} and then fine-tuning on the target environment. In these settings, our proposal brings in the advantage of enabling fine-tuning without reliance on depth annotations from the target environment, which would be costly or even prohibitive to collect. For all our tests we have used the DispNet-Corr1D \cite{Mayer_2016_CVPR} architecture, from now on shortened as \dispnet{}. Following the authors' guidelines \cite{Mayer_2016_CVPR}, we have trained a re-implementation of \dispnet{} on FlyingThings3D by the standard supervised L1 regression loss. Then, we have used these pre-trained weights as initialization for all the tests discussed hereinafter.
  
For our experiments we rely on the \kitti{} RAW \cite{KITTI_RAW} dataset, which features $\sim43K$ images with depth labels \cite{Uhrig2017THREEDV} converted into disparities by known camera parameters. Images are taken from stereo video sequences  concerning four diverse  environments, namely \emph{Road}, \emph{Residential}, \emph{Campus} and \emph{City}, containing 5674, 28067, 1149 and 8027 frames, respectively. Although all images come from driving scenarios, each environment shows peculiar traits that would lead a deep stereo model to gross errors without suitable fine-tuning. For example, \emph{City} and \emph{Residential} often depict road surrounded by buildings, while \emph{Road} mostly concerns highways and country roads where the most common objects are cars and vegetation. Using this data and extending the protocol introduced in \cite{Tonioni_2017_ICCV}, we wish to measure both \textit{target domain} performance, \ie, how the network performs on the target domain upon unsupervised adaptation without access to any groundtruth information,  as well as  \textit{similar domains} performance, \ie, how the network adapted unsupervisedly generalizes to unseen images from similar domains. To analyze both behaviours, we have alternatively used one of the environments as the training set to perform fine-tuning, then tested the resulting model on all the four environments. In fact, this allows for assessing   \textit{target domain performance} by testing on the environment used for unsupervised fine-tuning and \textit{similar domains performance} by testing on the other three. Since the environments amenable to perform fine-tuning are four, we can carry out 4-fold cross-validation in order to average performance figures. Hence, for each fold we average performance figures within an environment (\ie, across all of its frames), obtaining, thereby, four sets of measurements. Then, we compute  \textit{target domain} performance by averaging the scores dealing with the four training sets in the corresponding four folds and \textit{similar domains} performance by averaging across the other twelve scores.   

As for the per-frame performance figures, we compute both the Mean Average Error (MAE) and the percentage of pixels with disparity error larger than 3 (bad3) as suggested in \cite{KITTI_2012,KITTI_2015}. Due to image formats being different across the KITTI RAW dataset, we extract a central crop of size $320\times1216$ from each frame, which matches to the down-sampling factor of \dispnet{} and allows for validating almost all pixels with respect to the available groundtruth disparities. 

\subsubsection{Ablation Study}
\label{sssec:ablation}

Our previous work \cite{Tonioni_2017_ICCV} presented a detailed study on the impact of the hyper-parameters of the method for the depth-from-stereo networks, a similar discussion for depth-from-mono networks is reported in \autoref{sec:mono_kitti}.
Here, instead, we perform a more comprehensive ablation study aimed at answering the following questions: i) Can we simply use $D$ as noisy groundtruth without deploying $C$? ii) Is masking by $\tau$ really needed  or could we just use $C$ as a per-pixel weighting in $L_c$? iii) How important is the contribution of the additional loss terms $L_s$, $L_r$? iv) How is performance affected by the use of a learnable $\tau$?

To answer the above questions, we set AD as stereo algorithm, CCNN as confidence measure and run a set of experiments according to the cross validation protocol described in \autoref{ssec:stereo}. The resulting performance figures are reported in \autoref{table:ablation} as follows. Starting from the top row: (a) AD, \ie{} the stereo algorithm providing us with the noisy labels, (b)  \dispnet{} trained only on synthetic data (\ie{} the initial weights used for all the subsequent fine tuning), (c) \dispnet{} fine tuned to directly regress AD without deploying a confidence measure (\ie, minimization of the error plotted in \autoref{fig:tau_overlap}-(e)), (d) \dispnet{} fine tuned to minimize $L_c$ with $\tau=0$ (\ie, minimization fo the error plotted in \autoref{fig:tau_overlap}-(f) without explicit masking), (e-h) training to minimize different combinations of $L_c, L_s$ and $L_r$ with a fixed $\tau=0.8$, {(i) training with a learnable $\tau$ parameter or (j) by inferring it for each image with $\tau$Net}. The values for $\lambda_1, \lambda_2$ and $\tau$ (when fixed) are obtained by preliminary cross-validation with a methodology similar to that described in our previous work \cite{Tonioni_2017_ICCV}. Since rows (a) and (b) do not need any kind of fine tuning on  \kitti{} we report the same performances for both \textit{target} and \textit{similar domains}. 

To answer the question (i), we can compare results between rows (c) and (b). As expected, fine-tuning the network to directly regress the noisy measurements produced by AD is not a valid optimization strategy as it worsens the initial network performance both in the \textit{target domain} as well as in \textit{similar domains}. Interestingly, the network structure seems to behave as a regularizer and does not overfit too much to the noise in the labels, as testified by the huge performance gap between rows (c) and (a). 
To answer the question (ii), we can compare line (e) and (d), where the only difference is the value of $\tau$. The presence of $\tau=0.8$ in (e) helps improving performance by about $0.5\%$ bad3 while obtaining comparable performance in MAE. These results testify how masking out disparity measurements that are likely wrong yields better performance even though it increases the sparsity of the gradient signal actually deployed to update the model. A possible explanation for the close performance gap between (d) and (e) may be ascribed to the confidence maps produced by CCNN being highly bi-modal, with the vast majority of pixels carrying confidence scores equal to either 0 or 1. Therefore, even without applying a fixed threshold, many completely mistaken labels will see their contribution masked out during loss computation. 
To answer the question (iii) we can compare the performance reported in the last four rows. Adding $L_s$ in the optimization process does not improve  \textit{target domain} performance but slightly helps in  \textit{similar domains},  as clearly observable by comparing rows (f) and (e). The introduction of $L_r$, instead, seems more effective and results in improvement across all metrics, as shown by rows (g) and (e). Once again, larger improvements are obtained in case of unseen images from \textit{similar domains}. Furthermore, it is worth pointing out how our complete \textit{Adaptation} loss yields the best results, as vouched by the performance figures reported in  row  (h).
{Finally, to answer question (iv), we can compare rows (i) and (j) to row (h). Letting $\tau$ be a learnable parameter (i) may ease the overall training process by avoiding manual tuning or grid-search to find the optimal threshold while yielding only a slight performance decrease, \ie $+0.19\%$ and $+0.18\%$ bad3 in target and similar domains, respectively. Deploying the shallow $\tau$Net (j) to predict different thresholds places in between the two, showing improvements over learning a single $\tau$ but still not reaching the performance obtained through manual cross-validation.}

\autoref{fig:stereo_ablation} shows qualitative results related to the ablation study proposed in this subsection.  The top row depicts the reference image (a), the noisy disparities provided by AD (b) and the prediction produced by \dispnet{} trained only on synthetic data (c). The bottom row, instead, reports three different predictions obtained by the three adaptation approaches  referred to as \textit{Regression} (d), \textit{Weighted} (e) and \textit{Complete} (f) in \autoref{table:ablation}. By comparing (f) to (d) and (e) we can clearly verify that  our adaptation scheme can successfully mask out all the noise in the labels and learn only from good disparities. Moreover, we can perceive  the effectiveness of our adaptation approach  by comparing (f) to (c),  for example  by observing how it can significantly reduce the errors caused by the reflective surface on the right portion of the image, without at the same time introducing many artifacts, as unfortunately does happen in (c) and (d).    

\begin{table}[t]
\centering
\begin{tabular}{|l|cc|cc|}
\cline{2-5}
\multicolumn{1}{c|}{} & \multicolumn{2}{c|}{Target Domain} & \multicolumn{2}{c|}{Similar Domains} \\
\hline
Loss & bad3 & MAE & bad3 & MAE \\
\hline
(a) No Adaptation & 10.86 & 1.73 & 10.86 & 1.73 \\
(b) GT Tuned (K12/15) & 5.04 & 1.28 & 5.04 & 1.28 \\
(c) Godard et. al.\cite{godard2017unsupervised} & 4.01 & 1.07 & 4.20 & 1.09 \\
(d) Yinda et. al.\cite{zhang2018activestereonet} & 3.59 & 1.00 & 5.15 & 1.14 \\
\hline
(e) Tonioni et. al.\cite{Tonioni_2017_ICCV}-AD & 3.10 & 0.97 & 3.80 & 1.05\\
(f) \textit{Masked-AD+Smooth.} & 3.17 & 0.98 & 3.78 & 1.05 \\
(g) Tonioni et. al.\cite{Tonioni_2017_ICCV}-SGM & 2.73 &	0.93 & 3.71 & 1.09\\
(h) \textit{Masked-SGM+Smooth.} & 2.79 & 1.01 & 3.63 & 1.09 \\
\hline
(i) \textit{Adaptation-AD} ($\tau$=0.8) & 2.96 & 0.96 & 3.66 & 1.04 \\
(j) \textit{Learned Adaptation-AD} & 3.15 & 1.01 & 3.88 & 1.08 \\ 
(k) \textit{$\tau$Net-AD} & 3.15 & 0.99 & 3.83 & 1.07 \\
(l) \textit{Adaptation-SGM} ($\tau$=0.9) & \textbf{2.58} & \textbf{0.91} & 3.39 & \textbf{1.01} \\
(m) \textit{Learned Adaptation-SGM} & 2.84 & 0.99 & 3.75 & 1.07 \\ 
(n) \textit{$\tau$Net-SGM} & 2.71 & 0.97 & 3.54 & 1.05 \\
(o) \textit{Adaptation-AD-SGM} & 2.61 & 0.92 & \textbf{3.37} & \textbf{1.01}\\ 
(p) \textit{Learned Adaptation-AD-SGM} & 2.77 & 0.99 & 3.54 & 1.07 \\
(q) {\textit{$\tau$Net}-AD-SGM} & 2.79 & 0.97 & 3.67 & 1.07\\
\hline
\end{tabular}
\caption{Results obtained performing fine tuning of a pre-trained \dispnet{} network using different unsupervised strategy. All results are computed on the \kitti{} raw dataset using a 4-fold cross validation schema, best results highlighted in bold, our proposals in italic.}
\label{table:stereo}
\end{table}

\subsubsection{Comparison to other self-supervised losses}
\label{sss:adaptation}

We compare our proposal to other loss functions known in the literature that may be employed in order to fine-tune a deep stereo network without supervision. In particular, we consider two losses that, akin to ours, rely only on stereo frames to achieve a form of self-supervision: the appearance based re-projection and smoothness loss by Godard et al. \cite{godard2017unsupervised} and the local constraint normalization with window-based optimization loss of \cite{zhang2018activestereonet}. As the underlying principles and mechanisms are quite straightforward to reproduce, we have re-implemented the two losses following the authors' guidelines. 
Thus, we apply these alternative losses together with variants of our proposal, relying either on AD or SGM or both stereo algorithms, in order to fine-tune \dispnet{} upon pre-training on synthetic data. As an additional comparison, we also report results obtained by our previous loss formulation \cite{Tonioni_2017_ICCV} with both stereo algorithms. 
{When using AD together with SGM, we fuse the disparity maps according to the corresponding confidences.
For each pixel, we keep the disparity value with higher confidence among the two predictions. Then we obtain the corresponding confidence map as the pixel-wise max between those associated with the two algorithms.  
Finally, we consider all variants of our method: with a fixed $\tau=0.9$ (\textit{Adaptation}), a learned $\tau$  (\textit{Learned Adaptation}) or the output of $\tau$Net (\textit{$\tau$Net Adaptation}).}

Again, we follow the same 4-fold cross validation protocol as in \autoref{sec:experiments}. Results are reported in \autoref{table:stereo} alongside with the performance of the pre-trained  \dispnet{} model (No Adaptation) and those attainable by fine-tuning the pre-trained model by the LIDAR groundtruth available for the 400 frames of the \kitti{}2012 \cite{KITTI_2012} and \kitti{}2015 \cite{KITTI_2015} training sets (GT Tuned), \ie{} according to the standard training methodology adopted in the vast majority of works dealing with deep stereo. For the sake of fair comparison, all methods are evaluated based only on the disparity map predicted for the left frames of the stereo pairs and can not leverage additional external networks besides \dispnet{} (\ie, as for \cite{zhang2018activestereonet} we do not deploy also an external Invalidation Network).  

\begin{figure}
\includegraphics[width=0.48\textwidth]{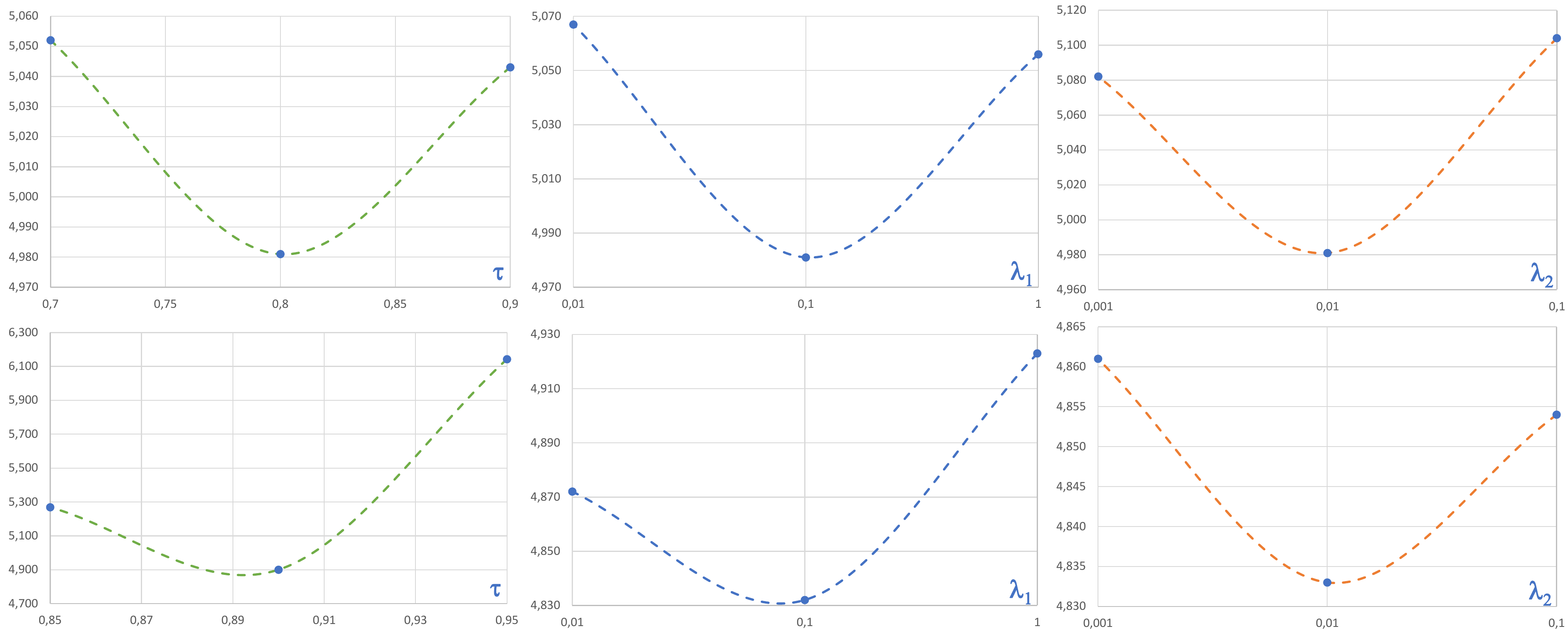} \\
\caption{Hyper-parameters study for unsupervised adaptation for \textit{monodepth} \cite{godard2017unsupervised}, VGG model. Top: AD algorithm, bottom: SGM. From left to right, RMSE achieved after 5 epochs of adaptation by varying respectively $\tau$, $\lambda_1$ and $\lambda_2$. Points are interpolated for visualisation purposes.}
\label{fig:mono_tuning}
\end{figure}

\autoref{table:stereo} shows that our proposal outperforms other approaches both in the \textit{target domain} as well as \textit{similar domain} experiments. In particular, \textit{Adaptation-SGM} (\textit{row k}) delivers the best performance on the target domain, with gain as large as $\sim1\%$ in the bad3 metric  with respect to the closest competitor known in literature beside our previous work, \ie, Yinda \etal{} at \textit{row d}. The improvement is less substantial in the MAE figure, though our proposal still consistently outperforms alternative approaches. 
We also point out how our original proposal  \cite{Tonioni_2017_ICCV}  (\textit{rows e} and \textit{d}) already outperforms competitors, which suggests the key component in our technique to be the confidence-guided loss. Yet, the novel \emph{Adaptation} scheme proposed in this paper further ameliorates performance significantly. {Moreover, from \textit{row e} to \textit{row h}  we compare the impact of the original smoothness proposed in \cite{Tonioni_2017_ICCV} to the edge-aware term introduced in this paper. In particular, while the former performs better on the target domain, the latter achieves lower errors when moving to similar domains, \eg{} $-0.08$ on bad3 when comparing \textit{row h} to \textit{row g}.}
By comparing \textit{Adaptation-AD} (\textit{row i}) to \textit{Adaptation-SGM} (\textit{row l}) we can verify how a more accurate stereo  algorithm (SGM vs AD) yields better performance. This can be ascribed to less noise in the disparities leading to a larger number of pixels scoring confidence $>\tau$  which, in turn, is conducive to denser and more accurate pseudo-groundtruths. 
{Using both stereo algorithms (\textit{Adaptation-AD-SGM} -- \textit{row o}) yields comparable performance to \textit{Adaptation-SGM} in both scenarios, with the best absolute perfomance in \textit{similar domains} and second best in \textit{target domain}. 
This behaviour might be explained considering that the errors of AD are not usually complementary to those of SGM due to the vast majority of pixels with low confidence with SGM corresponding to equally low confidence pixels with AD. Therefore, the fusion of the two algorithms does not add many new useful labels that our method may use, leading to a marginal improvement on similar domains compared to SGM alone ($-0.02\%$ bad3).  
Comparing the performance of methods with fixed $\tau$ (\ie, \textit{Adaptation} -- \textit{row i} and \textit{row l}) to those with $\tau$ as a learnable variable (\ie, \textit{Learned Adaptation} -- \textit{rows j, m, p}) we can see how the self-filtering strategy can ease the training process with a negligible loss in performance ($+0.2\%$ bad3 and $+0.06$ MAE), further reduced by estimating $\tau$ with a shallow network (\ie, \textit{$\tau$Net} -- \textit{rows k, n, q}).}

Finally, it is interesting to compare the performance achievable by fine-tuning without supervision on many data (\textit{rows} from \textit{e} to \textit{p}) to those achievable by fine-tuning with supervision on few similar data (\ie, \textit{GT Tuned} - \textit{row b}). The large performance margin in favour to most of unsupervised approaches indicates that training on much more data with a sub-optimal objective turns out not only easier and cheaper but also beneficial to performance with respect to training on few, perfectly annotated samples (\eg, $-1.65\%$ bad3 and $-0.27$ MAE by comparing  \textit{Adaptation-SGM} to \textit{GT Tuned}).


\begin{table*}[t]
\centering
\begin{tabular}{|c|c|cccc|ccc|}
\cline{5-8}
\multicolumn{4}{c}{} & \multicolumn{2}{|c|}{\cellcolor{blue!25}Lower is better}
 & \multicolumn{2}{c|}{\cellcolor{LightCyan}Higher is better} & \multicolumn{1}{c}{} \\
\hline
Supervision & Encoder & \cellcolor{blue!25} Abs Rel & \cellcolor{blue!25} Sq Rel & \cellcolor{blue!25} RMSE &\cellcolor{blue!25} RMSE log & \cellcolor{LightCyan} $\delta<$1.25 & \cellcolor{LightCyan} $\delta<1.25^2$ & \cellcolor{LightCyan}$\delta<1.25^3$\\
\hline
\hline
Godard et al. \cite{godard2017unsupervised} & VGG & 0.124 & 1.076 & 5.311 & 0.219 & 0.847 & 0.942 & 0.973 \\
\hline
Masked-AD & VGG & 0.119 & 0.989 & 4.981 & 0.207 & 0.859 & 0.950 & 0.977\\
Adaptation-AD & VGG & \textbf{0.118} & \textbf{0.976} & 5.009 & 0.206 & 0.859 & 0.949 & 0.977\\
{Learned Adaptation-AD} & VGG & 0.120 & 1.020 &  5.265 &  0.217 &  0.849 &  0.943 &  0.974\\
{$\tau$Net Adaptation-AD} & VGG & 0.119 & 0.976 &  5.096 &  0.213 &  0.854 &  0.946 &  0.974\\
\hline
Masked-SGM & VGG & 0.123 & 1.055 & 4.900 & 0.208 & 0.860 & 0.951 & 0.977 \\
{Adaptation-SGM} & VGG & 0.119 & 0.977 & \textbf{4.833} & 0.205 & 0.864 & \textbf{0.952} & \textbf{0.978}\\
{Learned Adaptation-SGM} & VGG & \textbf{0.118} & 1.015 &  5.166 &  0.213 &  0.854 &  0.947 &  0.975\\
{$\tau$Net Adaptation-SGM} & VGG & 0.126 & 1.213 &  5.113 &  0.214 &  0.859 &  0.953 &  0.976\\
\hline
Masked-AD-SGM & VGG & 0.122 & 1.049 & 4.975 & 0.207 & 0.857 & 0.950 & 0.976 \\
{Adaptation-AD-SGM} & VGG & 0.120 & 1.031 & 4.976 & \textbf{0.204} & \textbf{0.865} & \textbf{0.952} & \textbf{0.978} \\
{Learned Adaptation-AD-SGM} & VGG & 0.124 & 1.089 & 5.100 & 0.213 & 0.857 & 0.948 & 0.975 \\
{$\tau$Net Adaptation-AD-SGM} & VGG & 0.122 & 1.034 & 5.077 & 0.210 & 0.857 & 0.949 & 0.975 \\
\hline
\hline
Godard et al. \cite{godard2017unsupervised} & VGG+pp & 0.118 & 0.923 & 5.015 & 0.210 & 0.854 & 0.947 & 0.976 \\
\hline
Masked-AD & VGG+pp & \textbf{0.111} & 0.871 & 4.852 & 0.199 & 0.858 & 0.952 & 0.980 \\
{Adaptation-AD} & VGG+pp & \textbf{0.111} & 0.865 & 4.901 & 0.200 & 0.859 & 0.950 & 0.979\\
{Learned Adaptation-AD} & VGG+pp & 0.117 & 0.909 & 5.065 & 0.213 & 0.846 & 0.944 & 0.976\\
{$\tau$Net Adaptation-AD} & VGG+pp & 0.111 & 0.872 &  4.974 &  0.215 &  0.853 &  0.948 &  0.978\\
\hline
Masked-SGM & VGG+pp & 0.112 & 0.848 & 4.766 & 0.197 & 0.859 & 0.953 & \textbf{0.981} \\
{Adaptation-SGM} & VGG+pp & \textbf{0.111} & \textbf{0.840} & \textbf{4.744} & \textbf{0.197} & \textbf{0.862} &  \textbf{0.954} &  0.980\\
{Learned Adaptation-SGM} & VGG+pp & 0.114 & 0.890 &  4.961 & 0.207 & 0.853 & 0.948 & 0.978\\
{$\tau$Net Adaptation-SGM} & VGG+pp & 0.113 & 0.922 &  4.904 &  0.199 &  0.858 &  0.953 &  0.980\\
\hline
Masked-AD-SGM & VGG+pp & 0.114 & 0.915 & 4.909 & 0.199 & 0.859 & 0.953 & 0.980 \\
{Adaptation-AD-SGM} & VGG+pp & \textbf{0.111} & 0.902 & 4.863 & 0.199 & \textbf{0.862} & \textbf{0.954} & \textbf{0.981} \\
{Learned Adaptation-AD-SGM} & VGG+pp & 0.113 & 0.903 & 4.902 & 0.201 & 0.858 & 0.952 & 0.979 \\
{$\tau$Net Adaptation-AD-SGM} & VGG+pp & 0.112 & 0.892 & 4.913 & 0.200 & 0.859 & 0.952 & 0.979 \\
\hline
\hline
Godard et al. \cite{godard2017unsupervised} & ResNet50+pp & 0.114 & 0.898 & 4.935 & 0.206 & 0.861 & 0.949 & 0.976 \\
\hline
Masked-AD & ResNet50+pp & \textbf{0.109} & 0.867 & 4.810 & 0.197 & 0.866 & 0.953 & 0.979 \\
{Adaptation-AD} & ResNet50+pp & \textbf{0.109} & 0.867 & 4.852 & 0.196 & 0.866 & 0.954 & 0.978\\
{Learned Adaptation-AD} & ResNet50+pp & 0.110 & 0.864 & 4.953 & 0.195 & 0.858 & 0.948 & 0.976\\
{$\tau$Net Adaptation-AD} & ResNet50+pp & 0.109 & 0.863 &  4.927 &  0.204 &  0.858 &  0.948 &  0.976\\
\hline
Masked-SGM & ResNet50+pp & \textbf{0.109} & 0.837 & 4.703 & 0.194 & 0.867 & 0.955 & 0.980 \\
Adaptation-SGM & ResNet50+pp & \textbf{0.109} & \textbf{0.831} & \textbf{4.681} & \textbf{0.193} & 0.867 & \textbf{0.956} & \textbf{0.981} \\
{Learned Adaptation-SGM} & ResNet50+pp & 0.111 & 0.880 & 4.820 & 0.196 & 0.864 & 0.954 & 0.980\\
{$\tau$Net Adaptation-SGM} & ResNet50+pp & 0.109 & 0.858 &  4.794 &  0.196 &  0.865 &  0.954 &  0.979\\
\hline
Masked-AD-SGM & ResNet50+pp & 0.110 & 0.866 & 4.775 & 0.195 & 0.867 & 0.955 & 0.980 \\
{Adaptation-AD-SGM} & ResNet50+pp & 0.110 & 0.891 & 4.809 & 0.196 & \textbf{0.868} & \textbf{0.956} & \textbf{0.981} \\
{Learned Adaptation-AD-SGM} & ResNet50+pp & 0.110 & 0.879 & 4.838 & 0.198 & 0.864 & 0.953 & 0.979 \\
{$\tau$Net Adaptation-AD-SGM} & ResNet50+pp & 0.110 & 0.872 & 4.837 & 0.198 & 0.863 & 0.953 & 0.979 \\

\hline
\end{tabular}
\caption{Experimental results on the \kitti{} dataset \cite{KITTI_RAW} on the data split proposed by Eigen et al. \cite{eigen2014depth}. On even conditions, the proposed adaptation scheme outperforms the supervision by Godard et al. \cite{godard2017unsupervised}.}
\label{table:mono}
\end{table*}

{\subsection{Depth-from-Mono}\label{sec:mono}}

To investigate the application of our approach to depth prediction from a single image, we run experiments based on the popular depth-from-mono system developed by Godard et al. \cite{godard2017unsupervised}. This choice is driven by two main factors i) despite a large number of works in this field \cite{zhou2017unsupervised,mahjourian2018unsupervised,wang2018unsupervised,yin2018geonet}, it still represents one of the most effective solutions for unsupervised depth-from-mono and ii) the image reconstruction loss proposed by Godard et al. represent the main competitor to our approach. Thus the comparison to \cite{godard2017unsupervised} turns out the ideal test bench for our proposal.

The network proposed in \cite{godard2017unsupervised}, referred to here as \textit{monodepth}, consists in a DispNet-like architecture featuring a backbone encoder followed by  a decoder to restore the original input resolution and predict the final depth map. In \cite{godard2017unsupervised}, both VGG \cite{simonyan2014very} and ResNet50 \cite{he2016deep} were tested as encoders.
The output is provided as disparity (\eg, inverse depth), and used at training time to warp the stereo images. This also eases the use of our unsupervised adaptation technique, that could be deployed anyway also in case of architectures directly predicting depth by simply converting our disparity labels based on known camera parameters.
Moreover, in \cite{godard2017unsupervised} a post-processing step is proposed to deal with occlusions and artifacts inherited from stereo supervision, by producing both normal and flipped depth maps and combining them. We will run experiments with and without this optional step, referred to as '+pp'.

We start from the TensorFlow codebase provided by the authors of \cite{godard2017unsupervised}, adding our proposal therein and running experiments within the same framework to ensure perfectly fair test conditions.

\subsubsection{Evaluation protocol}

We follow exactly the same protocol as reported in \cite{godard2017unsupervised}. In particular, the \kitti{} raw dataset \cite{KITTI_RAW} is split into a training set and an evaluation set according to the guidelines by Eigen et al. \cite{eigen2014depth}. 
Unlike the adopted stereo evaluation protocol \cite{Uhrig2017THREEDV},  raw LiDAR measurements are usually assumed as groundtruth in the depth-from-mono literature despite their being sparse and noisy. Nonetheless, we adhere to the standard depth-from-mono evaluation protocol to ensure consistency with existing literature and enable a fair comparison with respect to \cite{godard2017unsupervised}.

Several works in this field \cite{zhou2017unsupervised,godard2017unsupervised,yin2018geonet} deploy  pre-training on the CityScapes dataset \cite{cordts2016cityscapes} before fine-tuning on the \kitti{}  training split \cite{eigen2014depth}, \cite{KITTI_RAW}. Indeed, training only on \kitti{} leads to inferior accuracy due to the fewer training images, whilst training only on CityScapes let the networks predicts depth maps of reasonable visual quality but totally wrong in terms of the actual depth values. This scenario, thus, points out again how a domain shift severely affects the accuracy of depth-from-images networks, \ie{} exactly the issue we aim to address by the general domain adaptation framework proposed in this paper. Therefore, to assess the effectiveness of our proposal also in depth-from-mono settings, we will start from models pre-trained on CityScapes in order to adapt them to \kitti{}. In particular, relying on the very same models pre-trained on CityScapes we compare the results attained on the \kitti{} test split by performing fine-tuning on the  \kitti{} train split by either our approach or the reconstruction loss proposed in \cite{godard2017unsupervised}. As for our method, we use the same stereo algorithms (AD and SGM), confidence measure (CCNN) and hyper-parameter settings as in depth-from-stereo experiments. Coherently to \cite{godard2017unsupervised}, we used the Adam optimizer and found that, while our competitor needs to run 50 epochs of training on \kitti{}, our method reaches convergence after only 5 epochs with a fixed learning rate of 0.001, thus resulting in faster and, as we shall see in the next section, more effective adaptation.

\begin{table*}[t]
\centering
\begin{tabular}{|l|c|cccc|ccc|}
\cline{5-8}
\multicolumn{4}{c}{} & \multicolumn{2}{|c|}{\cellcolor{blue!25}Lower is better}
 & \multicolumn{2}{c|}{\cellcolor{LightCyan}Higher is better} & \multicolumn{1}{c}{} \\
\hline
Configuration & Encoder & \cellcolor{blue!25} Abs Rel & \cellcolor{blue!25} Sq Rel & \cellcolor{blue!25} RMSE &\cellcolor{blue!25} RMSE log & \cellcolor{LightCyan} $\delta<$1.25 & \cellcolor{LightCyan} $\delta<1.25^2$ & \cellcolor{LightCyan}$\delta<1.25^3$\\
\hline
Regression-AD & VGG+pp & 0.209 & 2.121 & 7.788 & 0.402 & 0.639 & 0.818 & 0.900 \\
Weighted-AD & VGG+pp & 0.124 & 1.010 & 5.446 & 0.236 & 0.825 & 0.932 & 0.968 \\
Masked-AD & VGG+pp & 0.111 & 0.871 & 4.852 & 0.199 & 0.858 & 0.952 & 0.980\\
\hline
Regression-SGM & VGG+pp & 0.136 & 1.697 & 5.540 & 0.220 & 0.848 & 0.942 & 0.973 \\
Weighted-SGM & VGG+pp & 0.117 & 0.983 & 4.987 & 0.202 & 0.857 & 0.951 & 0.979 \\
Masked-SGM & VGG+pp & 0.112 & 0.848 & 4.766 & 0.197 & 0.859 & 0.953 & 0.981\\
\hline
\hline
Regression-AD & ResNet50+pp & 0.230 & 3.240 & 8.361 & 0.418 & 0.624 & 0.806 & 0.893 \\
Weighted-AD & ResNet50+pp & 0.120 & 0.952 & 5.288 & 0.225 & 0.836 & 0.937 & 0.971 \\
Masked-AD & ResNet50+pp & 0.109 & 0.867 & 4.810 & 0.197 & 0.866 & 0.953 & 0.979\\
\hline
Regression-SGM & ResNet50+pp & 0.129 & 1.456 & 5.385 & 0.214 & 0.854 & 0.943 & 0.973 \\
Weighted-SGM & ResNet50+pp & 0.115 & 0.966 & 4.925 & 0.199 & 0.863 & 0.952 & 0.979 \\
Masked-SGM & ResNet50+pp & 0.109 & 0.837 & 4.703 & 0.194 & 0.867 & 0.955 & 0.980\\
\hline
\end{tabular}
\caption{Ablation experiments on the \kitti{} dataset \cite{KITTI_RAW} on the data split proposed by Eigen et al. \cite{eigen2014depth}.}
\label{table:mono_ablation}
\end{table*}

\subsubsection{Results on \kitti{}}\label{sec:mono_kitti}

We discuss here the outcomes of our experiments on the \kitti{} RAW dataset \cite{KITTI_RAW}. In particular, we report the standard error metrics, \ie{}  Absolute Relative error (Abs Rel), Square Relative error (Sq Rel), Root Mean Square Error (RMSE), logarithmic RMSE and the $\delta$ accuracy score computed as:

\begin{equation}
    \delta = \tilde{D}_{i,j}\% : \max (\frac{\tilde{D}_{i,j}}{D_{i,j}},\frac{\tilde{D}_{i,j}}{\tilde{D}_{i,j}})  < th
\end{equation}
{Hyper-parameters $\tau, \lambda_1$ and $\lambda_2$ were manually tuned to obtain the best accuracy. Figure \ref{fig:mono_tuning} reports how the RMSE metric behaves by varying each of the three parameters while adapting the VGG model on either AD (top) or SGM (bottom). We found configurations $\tau=0.8,\lambda_1=0.1,\lambda_2=0.01$ and $\tau=0.9,\lambda_1=0.1,\lambda_2=0.01$ to be the best  for AD and SGM, respectively.}

Table \ref{table:mono} reports a detailed comparison between the self-supervised loss proposed in \cite{godard2017unsupervised} and our proposal in the aforementioned configurations \textit{Masked}, \textit{Adaptation}, \textit{Learned} and \textit{$\tau$Net Adaptation}, all applied to the same \textit{monodepth} model pre-trained on CityScapes by the authors \cite{godard2017unsupervised}. From top to bottom, we show the results dealing with VGG, VGG using post-processing step (+pp) and ResNet50 +pp models. {The best metrics across the different configurations on a single model are higlighted in bold.

Starting from the basic VGG on top, we can observe that adapting by either AD, SGM or both combined with the Masked configuration alone leads to better performance with respect to using the image reconstruction loss proposed in \cite{godard2017unsupervised}. In general, adapting by SGM yields superior results, outperforming the model based on AD in nearly all metrics. Applying our full adaptation scheme yields further improvements in almost all metrics with respect to the results achieved by the confidence guided loss alone. Contextually, we point out that combining AD and SGM achieves similar performance as observed for stereo experiments, leading to the best $\delta < 1.25^2$ and $\delta < 1.25^3$ together with \textit{Adaptation-SGM} and achieving alone the best $\delta < 1.25$ score. {Moreover, the \textit{Learned Adaptation} scheme always achieves slightly worse results compared to a hand-tuned threshold $\tau$, with \textit{$\tau$Net Adaptation} placing in between the two alternatives. Nonetheless, all adaptation proposals turn out more accurate than the loss by Godard \etal{} \cite{godard2017unsupervised}.}

This finding is confirmed when applying the post-processing step (i.e., VGG+pp), as our adaptation approach outperforms \cite{godard2017unsupervised} under all evaluation metrics. Moreover, VGG+pp networks optimized by variants of our technique can deliver better results than using a ResNet+pp network trained according to the image reconstruction loss of  \cite{godard2017unsupervised}, despite the large difference in complexity between the two networks (VGG features about 31 millions learnable parameters, ResNet50 about 57 millions). In this case, \textit{}{Adaptation-SGM} consistently achieves the best results on most metrics, except for $\delta < 1.25^3$ where \textit{Masked-SGM} and \textit{Adaptation-AD-SGM} slightly outperforms it. {Again, learning $\tau$, by either the \textit{Learned} or \textit{$\tau$Net} strategy, leads to better results than Godard \etal{} on most metrics, although slightly reducing the effectiveness of our adaptation scheme.}

Moving to ResNet50+pp model, the margin turns out even higher. We highlight once more how all the variants of our technique consistently outperforms Godard \etal{} on almost all cases. Similarly to VGG+pp, the lowest error metrics are achieved by Adaptation-SGM, while the highest $\delta < 1.25^2$ and $\delta < 1.25^3$ are sourced by both Adaptation-SGM and Adaptation-AD-SGM, being finally $\delta < 1.25$ better for the latter strategy thanks to the combination of the two stereo algorithms. 
{Finally, determining $\tau$ by either the \textit{Learned} or \textit{$\tau$Net} strategy yields, again, to minor drops in almost all metrics. Thus, it may represent a practical alternative to explicit hand-tuning of $\tau$.}}

\begin{figure*}
\begin{tabular}{ccc}
\includegraphics[width=0.30\textwidth]{images/eigen_045.jpg} &
\begin{overpic}[width=0.30\textwidth]{images/ad.jpg}
\put (2,25) {$\displaystyle\textcolor{white}{\textbf{bad3: 38.12}}$}
\end{overpic} &
\begin{overpic}[width=0.30\textwidth]{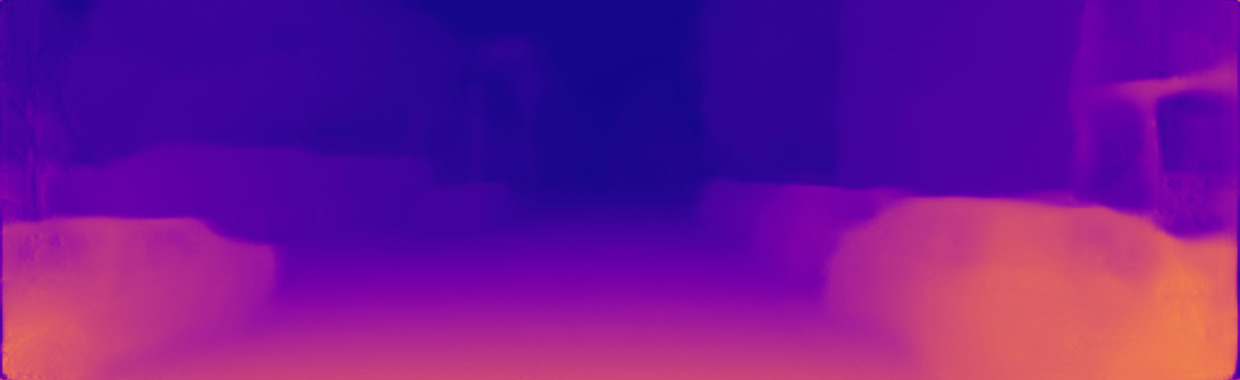}
\put (2,25) {$\displaystyle\textcolor{white}{\textbf{Abs Rel: 0.602}}$}
\end{overpic} \\
(a) &
(b) &
(c) \\
\begin{overpic}[width=0.30\textwidth]{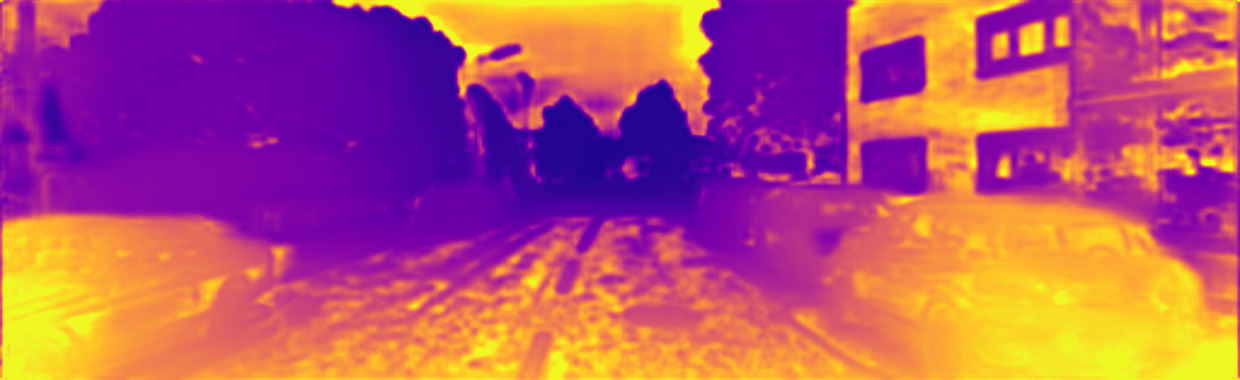}
\put (2,25) {$\displaystyle\textcolor{white}{\textbf{Abs Rel: 0.203}}$}
\end{overpic} &
\begin{overpic}[width=0.30\textwidth]{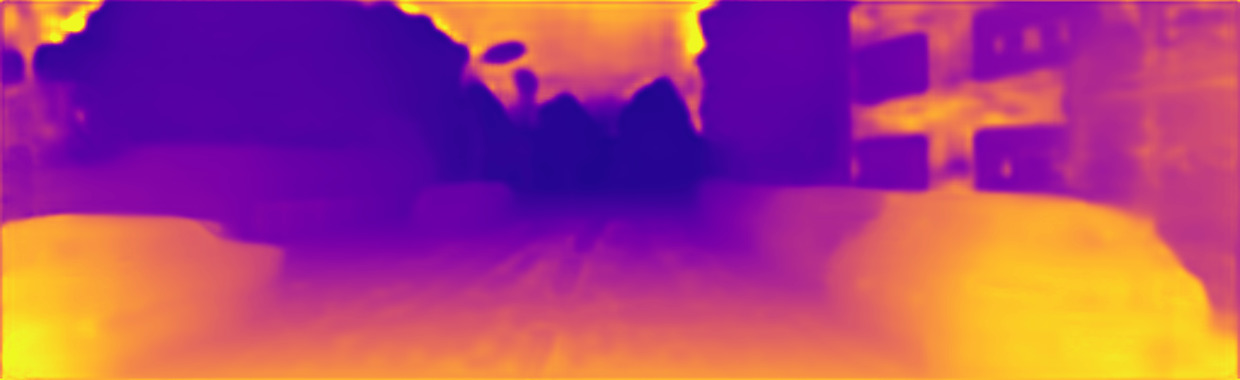}
\put (2,25) {$\displaystyle\textcolor{white}{\textbf{Abs Rel: 0.120}}$}
\end{overpic} & 
\begin{overpic}[width=0.30\textwidth]{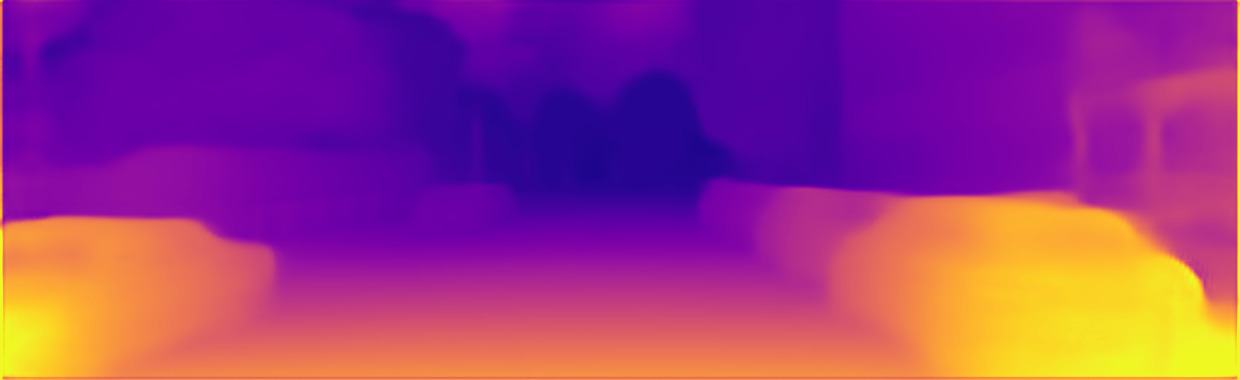}
\put (2,25) {$\displaystyle\textcolor{white}{\textbf{Abs Rel: 0.098}}$}
\end{overpic} \\
(d) &
(e) &
(f) \\
\end{tabular}
\caption{Ablation experiments: adaptation of \textit{monodepth} (VGG encoder) using AD algorithm. a) input image from \kitti{} b) result from AD algorithm c) result before adaptation d) adapting with stereo algorithm only e) using confidence to weight the loss function f) running full adaptation.}
\label{fig:mono_ablation}
\end{figure*}

\subsubsection{Ablation experiments}

Similarly to the stereo settings previously addressed in Table \ref{table:ablation}, we report here an ablation study aimed at establishing the relative importance of the key ingredients deployed in our framework. Table \ref{table:mono_ablation} collects the results obtained in this evaluation. We comment about four main experiments, dealing with running our method with both AD and SGM in order to adapt VGG and ResNet50. The post-processing step is enabled in all tests, thereby solving most issues near occlusions and left border and highlighting how the full confidence-guided loss ameliorates results in many regions of the images where post-processing cannot operate. Three setups are considered in descending order in the Table for each of the four experiments: i) adaptation by minimization of the L1 loss with respect to the disparity maps estimated by the stereo algorithm (AD or SGM) "as is" (\textit{Regression}) ii) adaptation by weighting the L1 loss with per-pixel confidence scores (\textit{Weighted}) iii) full confidence-guided loss using threshold $\tau$ (\textit{Masked}). We turn off additional terms to focus on the different key factors of the confidence-guided loss.
In all experiments, we can notice how using the disparity labels alone leads to poor results, in particular when adapting the model by the AD algorithm, which is much more prone to outliers. This further highlights how, in our framework, deploying the confidence measure is crucial to avoid the impact of the wrong disparities possibly computed by the stereo algorithms.
Formulating the confidence-guided loss as a simple weighting between confidence scores and loss signals reduces the impact of the outliers, but does not completely removes it as they can still contribute to the entire loss function with a lower weight and thus may lead, as reported, to worse performance.
To better perceive this effect, \autoref{fig:mono_ablation} shows some qualitative results obtained by the three ablated configurations reported in the Table. In particular, we point out how on (c) the results from the original model trained on different environments look good qualitatively, but the range of the predicted depth values is totally wrong (Abs Rel of 0.620). We can observe how ablated configurations of our technique (d-e) do yield gradual improvements, whereas the full adaptation scheme (f) greatly ameliorates the quality of the estimated depth maps, \ie{} so as to bring the error down to  0.098 Abs Rel. 

\begin{figure}
\setlength\tabcolsep{0.5pt}
\begin{tabular}{ccc}
Input & No Adaption & Adaptation \\
\includegraphics[width=0.16\textwidth]{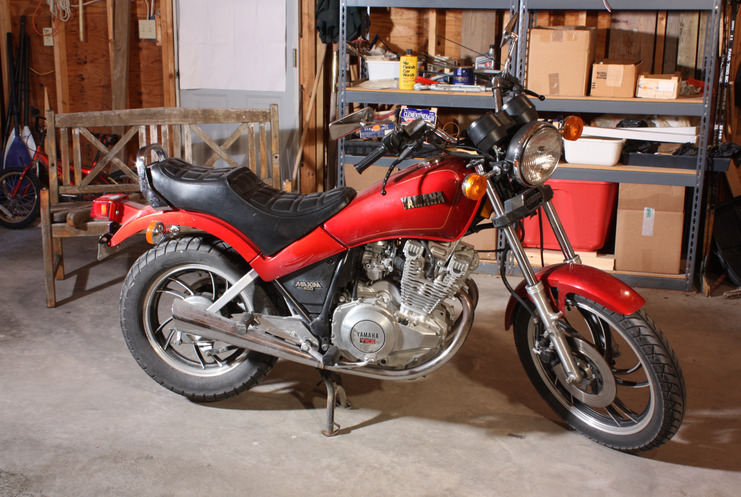} &
\begin{overpic}[width=0.16\textwidth]{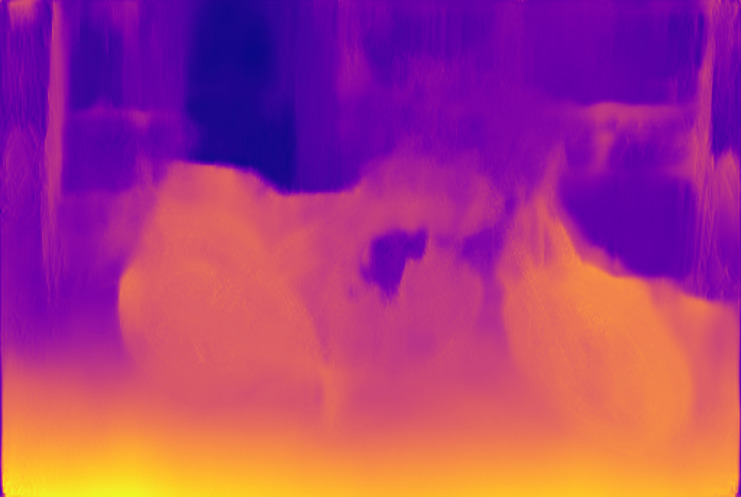} 
\put (2,55) {$\displaystyle\textcolor{white}{\textbf{Abs Rel: 0.6827}}$}
\end{overpic} &
\begin{overpic}[width=0.16\textwidth]{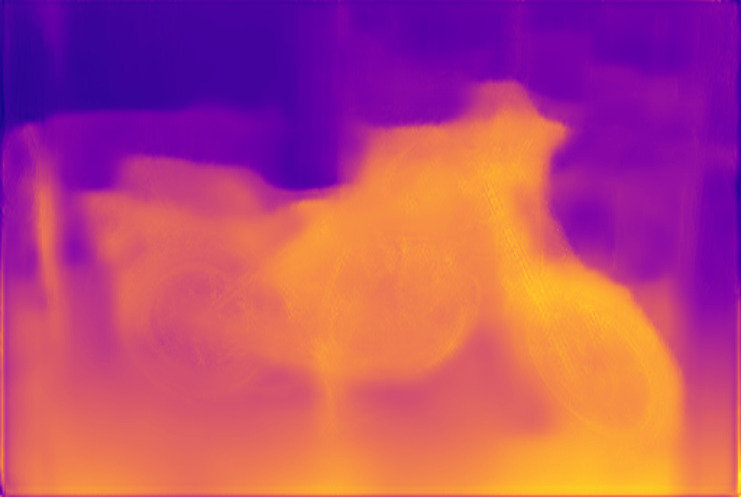} 
\put (2,55) {$\displaystyle\textcolor{white}{\textbf{Abs Rel: 0.1797}}$}
\end{overpic}
\\
\includegraphics[width=0.16\textwidth]{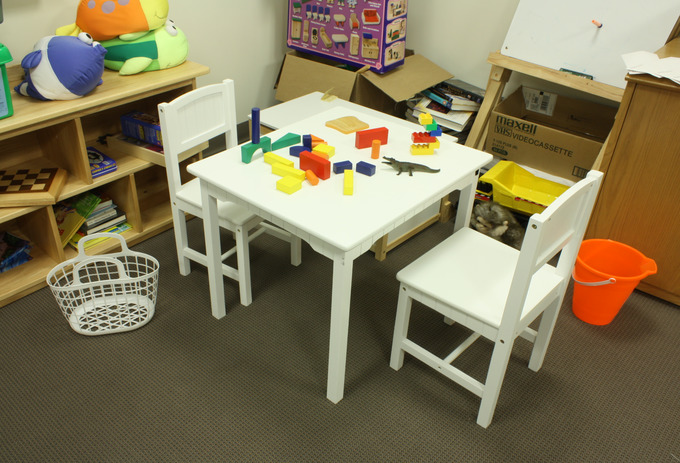} &
\begin{overpic}[width=0.16\textwidth]{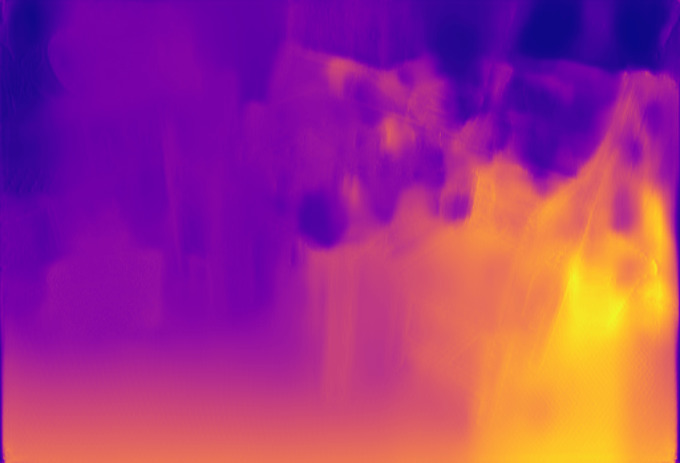} 
\put (2,57) {$\displaystyle\textcolor{white}{\textbf{Abs Rel: 0.9996}}$}
\end{overpic} &
\begin{overpic}[width=0.16\textwidth]{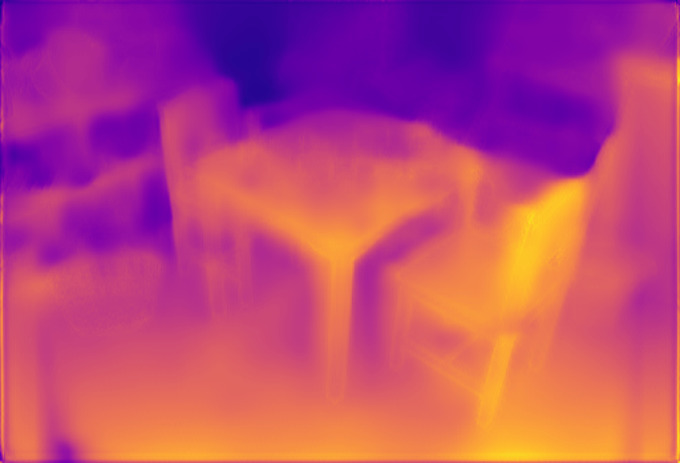} 
\put (2,57) {$\displaystyle\textcolor{white}{\textbf{Abs Rel: 0.1271}}$}
\end{overpic} \\
\\
\hline
\\
\includegraphics[width=0.16\textwidth]{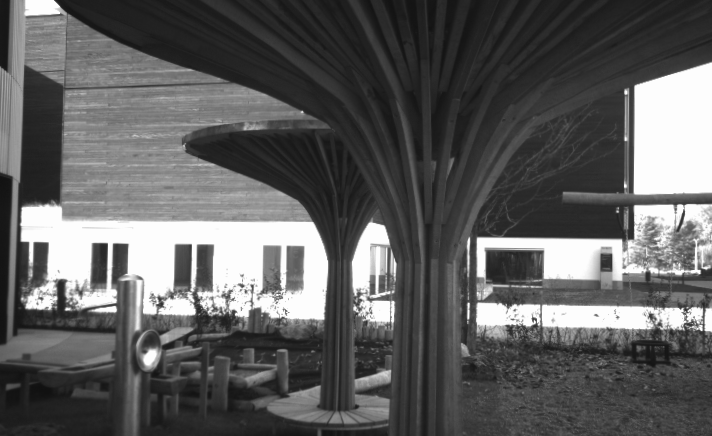} &
\begin{overpic}[width=0.16\textwidth]{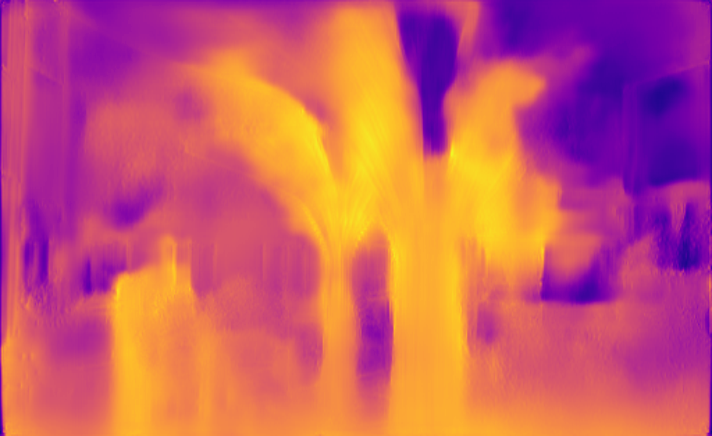} 
\put (2,48) {$\displaystyle\textcolor{white}{\textbf{Abs Rel: 0.6466}}$}
\end{overpic} &
\begin{overpic}[width=0.16\textwidth]{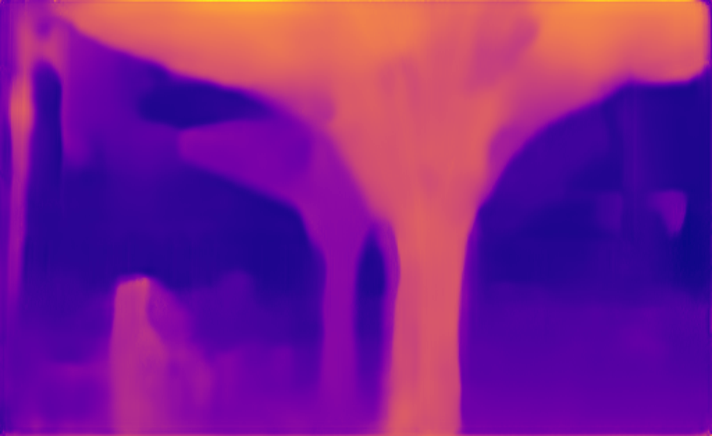} 
\put (2,48) {$\displaystyle\textcolor{white}{\textbf{Abs Rel: 0.1655}}$}
\end{overpic} \\
\includegraphics[width=0.16\textwidth]{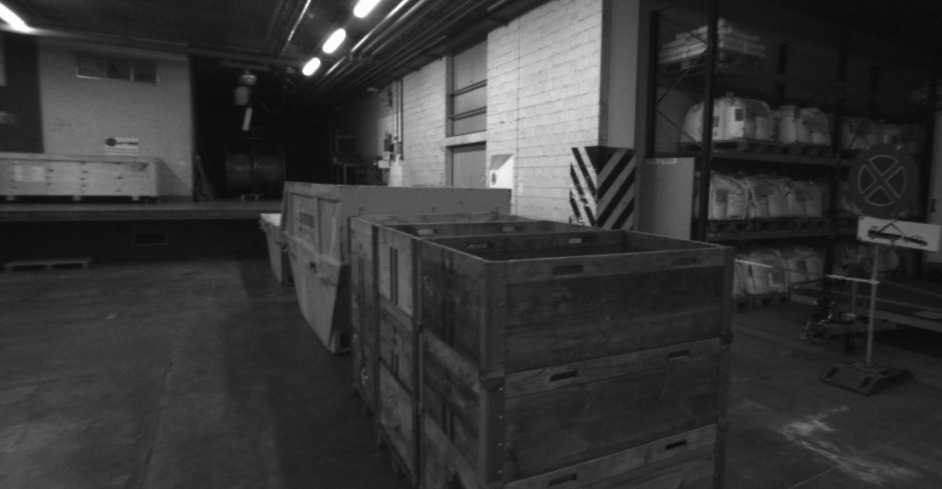} &
\begin{overpic}[width=0.16\textwidth]{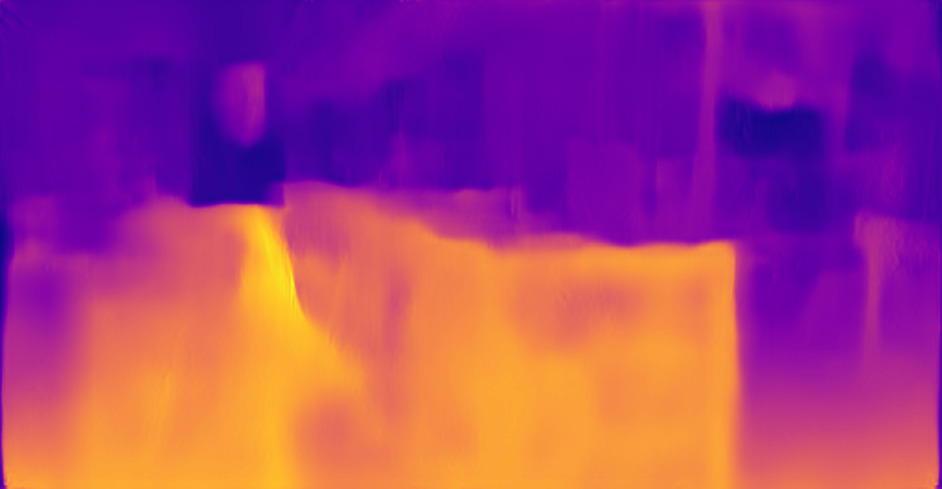} 
\put (2,42) {$\displaystyle\textcolor{white}{\textbf{Abs Rel: 0.6740}}$}
\end{overpic} &
\begin{overpic}[width=0.16\textwidth]{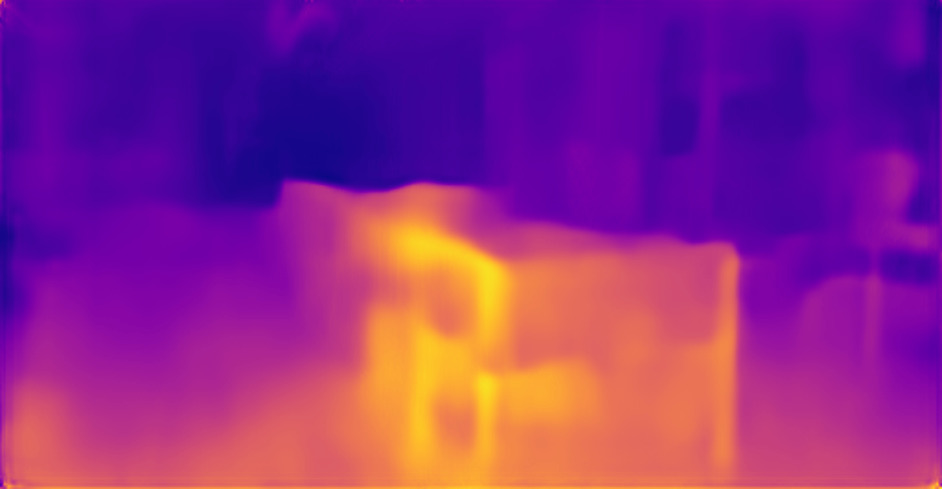} 
\put (2,42) {$\displaystyle\textcolor{white}{\textbf{Abs Rel: 0.1827}}$}
\end{overpic} \\
\end{tabular}
\caption{Adaptation results for depth-from-mono on Middlebury v3 \cite{MIDDLEBURY_2014} (top) ETH3D dataset \cite{schoeps2017cvpr} (bottom). From left to right: input (left) image, depth maps from network before adaptation and after fine tuning with our adaptation technique. The absolute error rate is overimposed on each depth map.}
\label{fig:qualitative_mono}
\end{figure}

{
\subsection{Analysis of $\tau$ convergence}
\label{ssec:additional}
\begin{figure}
    \centering
    \includegraphics[width=\linewidth]{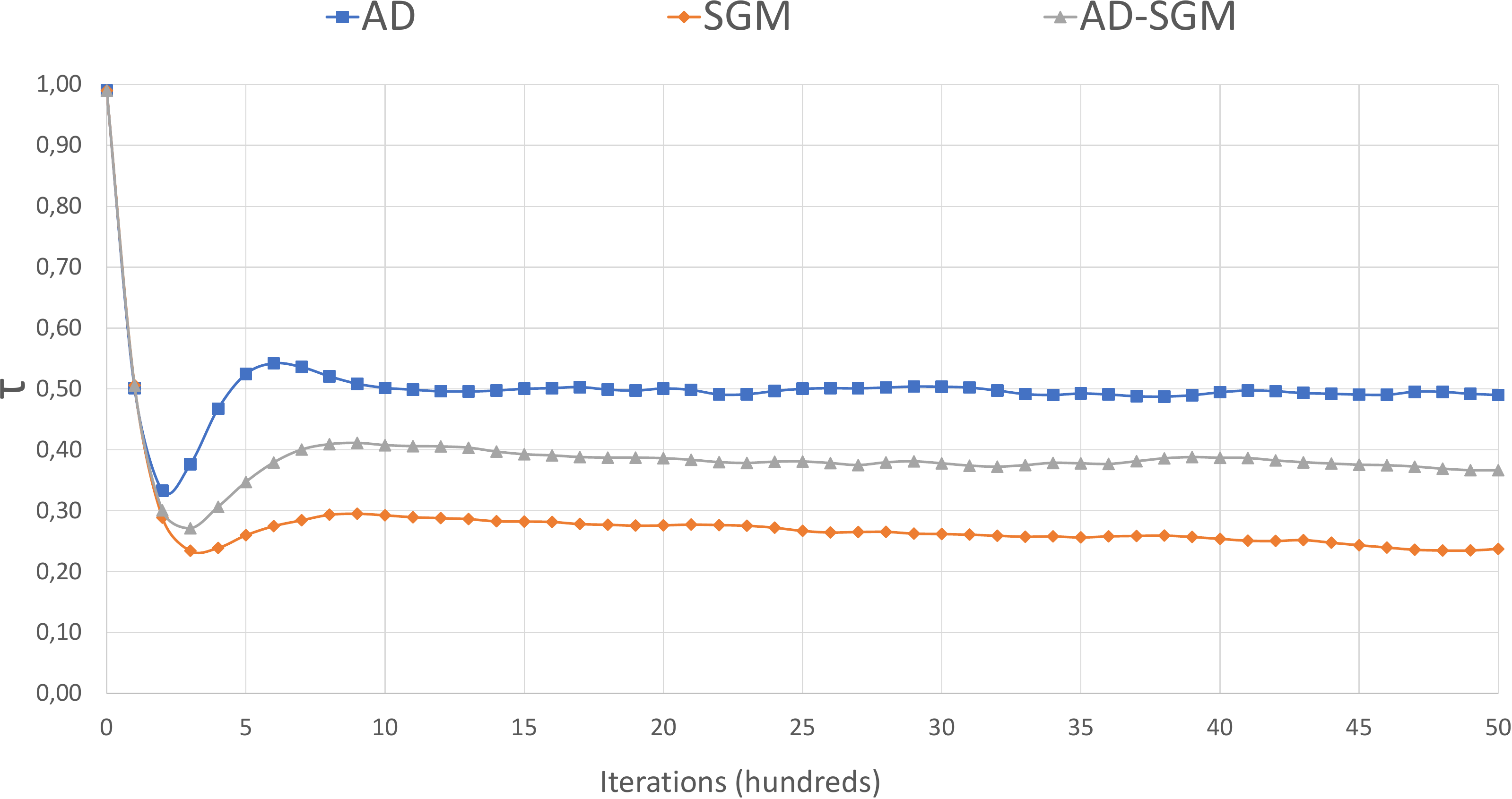}
    \caption{Learned values of $\tau$ across three different training using different stereo algorithms and CCNN as confidence measure. }
    \label{fig:plot_tau}
\end{figure}
To get insights on which values are automatically selected for $\tau$ using the learned adaptation scheme presented in \autoref{ssec:learning_tau}, we plot in \autoref{fig:plot_tau} the value of the variable across $5000$ training iterations  using either AD, SGM or the mixed stereo dataset {and directly optimizing $\tau$ as a learnable parameter.
In all the three runs, $\tau$ was initialized to $0.99$ and then updated by gradient descent  along with the other parameters. 
Similar behaviours are observed adapting both stereo and mono models therefore we report only the former. }

The plot shows how across the three runs the value of $\tau$ starts to stabilize around $1000$ iterations after an initial drop and subsequent rebound in the first $500$. This occurs when the disparity loss surpasses the penalty term after several outliers have been included, thus preventing $\tau$ to decrease further. 
Overall the behaviour of $\tau$ resembles a \emph{curriculum learning}\cite{bengio2009curriculum} schedule. At the beginning a high $\tau$ value filters out most low-confidence pixels while keeping only high confidence ones, \ie{} an easier regression task to learn. 
Then, $\tau$ starts decreasing, thereby considering more pixels, as well as noise, in the loss estimation process, \ie{} the optimization task for the network becomes harder. 
In the end, the value of $\tau$ stabilizes to a reasonable threshold for both the considered stereo algorithms, with AD ending up to a higher value due to its higher amount of outliers. Consistently, the learned $\tau$ for AD-SGM is higher than that of SGM alone, suggesting how the fusion strategy introduce errors from AD within the  SGM predictions.
{Concerning $\tau$Net we observed empirically that the predicted values of $\tau$, on average, exhibit a similar behaviour with a slightly higher variance due to $\tau$Net being a function of the current input and not a global threshold.} 

Compared to the fixed $\tau$,  both learning strategy produce lower threshold, thus introducing more outliers during adaptation. Nevertheless, as hand-tuning by cross-validation is unlikely to happen in a real scenario without any available groundtruth,  learning  $\tau$ by the proposed techniques represents an effective strategy.
}

\subsection{Qualitative Results}

\begin{figure}
\setlength\tabcolsep{0.5pt}
\begin{tabular}{ccc}
Input & No Adaption & Adaptation \\
\includegraphics[width=0.16\textwidth]{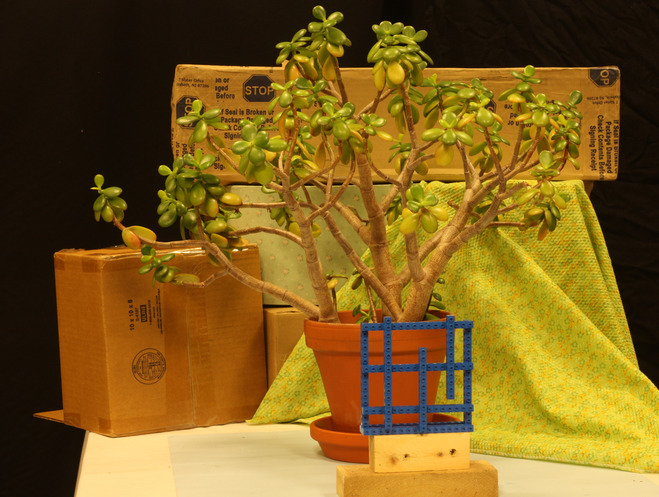} &
\begin{overpic}[width=0.16\textwidth]{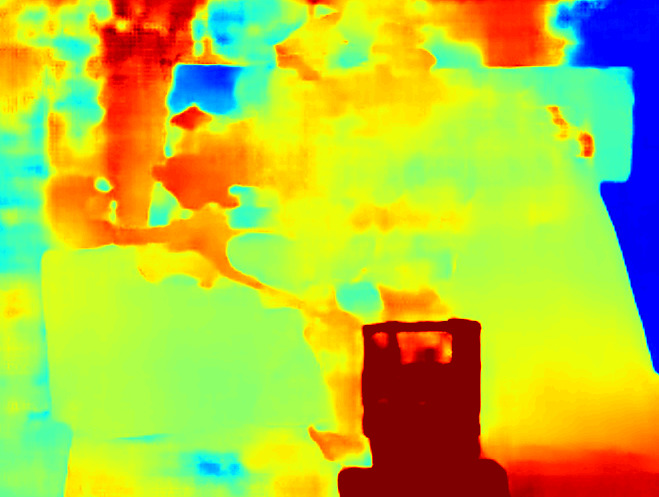} 
\put (2,65) {$\displaystyle\textcolor{white}{\textbf{Bad1: 88.94\%}}$}
\end{overpic} &
\begin{overpic}[width=0.16\textwidth]{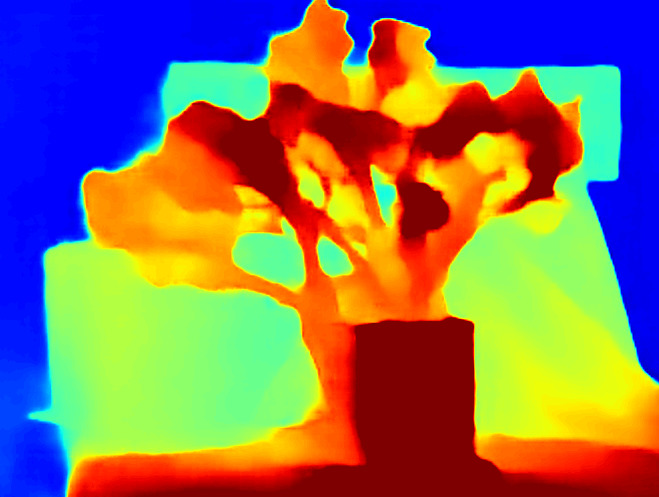} 
\put (2,65) {$\displaystyle\textcolor{white}{\textbf{Bad1: 47.28\%}}$}
\end{overpic}
\\
\includegraphics[width=0.16\textwidth]{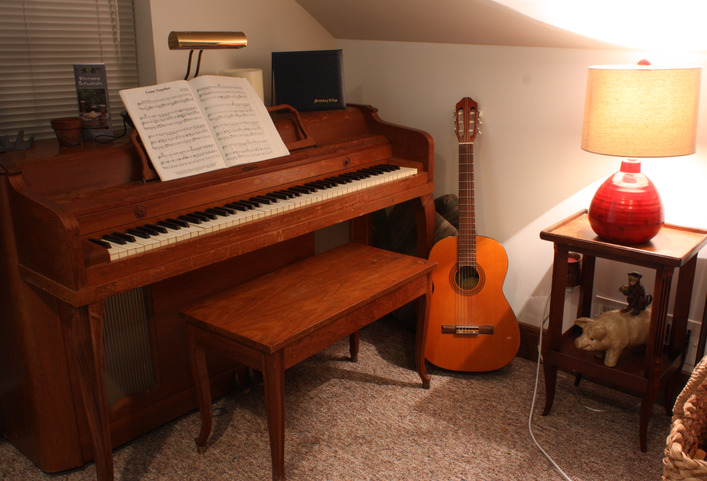} &
\begin{overpic}[width=0.16\textwidth]{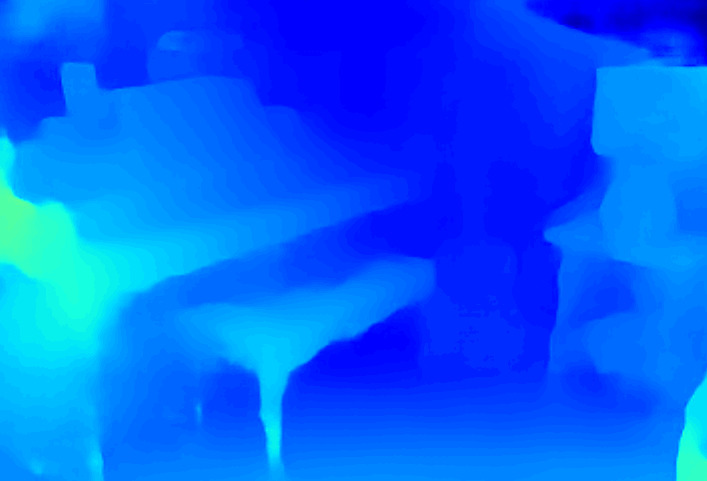} 
\put (2,57) {$\displaystyle\textcolor{white}{\textbf{Bad1: 40.54\%}}$}
\end{overpic} &
\begin{overpic}[width=0.16\textwidth]{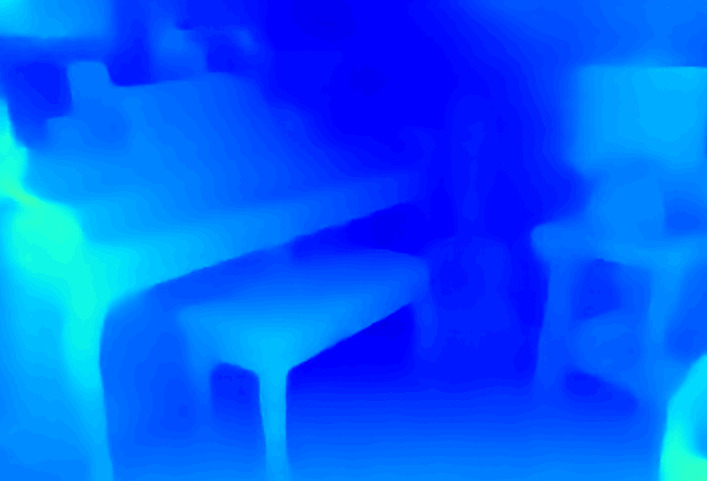} 
\put (2,57) {$\displaystyle\textcolor{white}{\textbf{Bad1: 21.54\%}}$}
\end{overpic} \\
\\
\hline
\\
\includegraphics[width=0.16\textwidth]{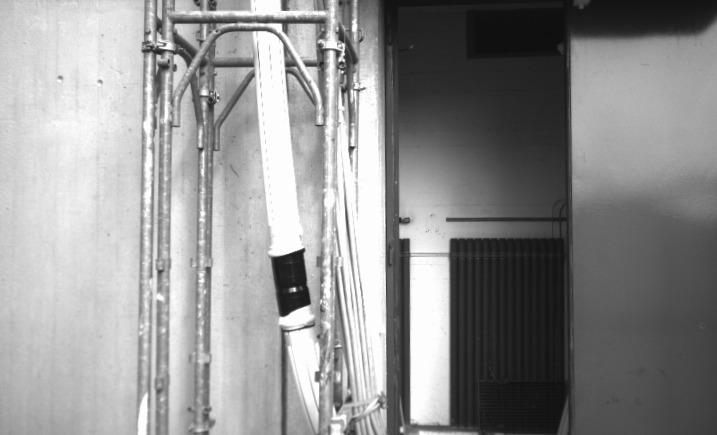} &
\begin{overpic}[width=0.16\textwidth]{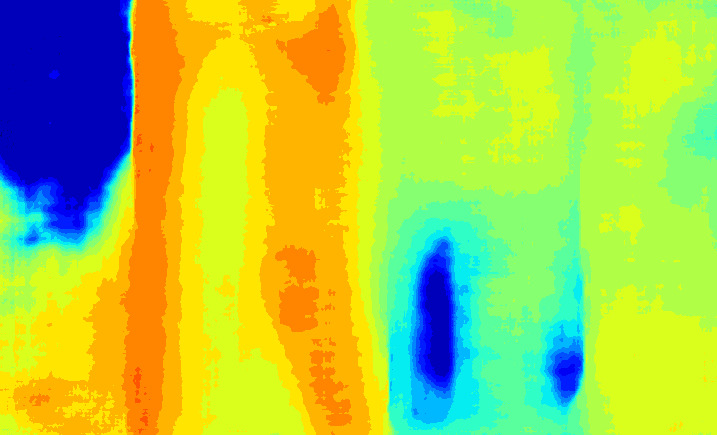} 
\put (2,48) {$\displaystyle\textcolor{white}{\textbf{Bad1: 63.30\%}}$}
\end{overpic} &
\begin{overpic}[width=0.16\textwidth]{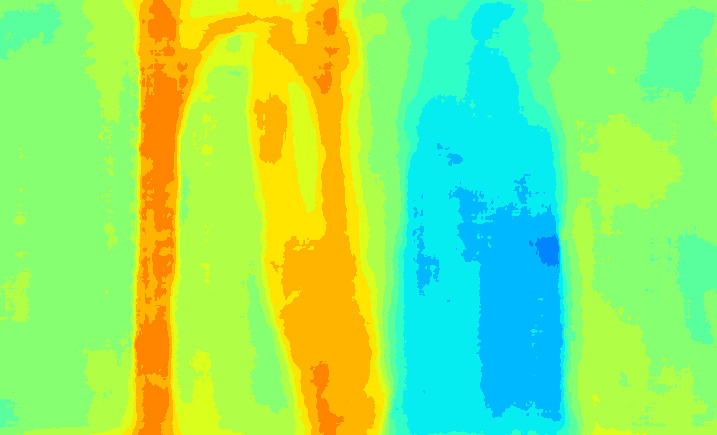} 
\put (2,48) {$\displaystyle\textcolor{white}{\textbf{Bad1: 7.29\%}}$}
\end{overpic} \\
\includegraphics[width=0.16\textwidth]{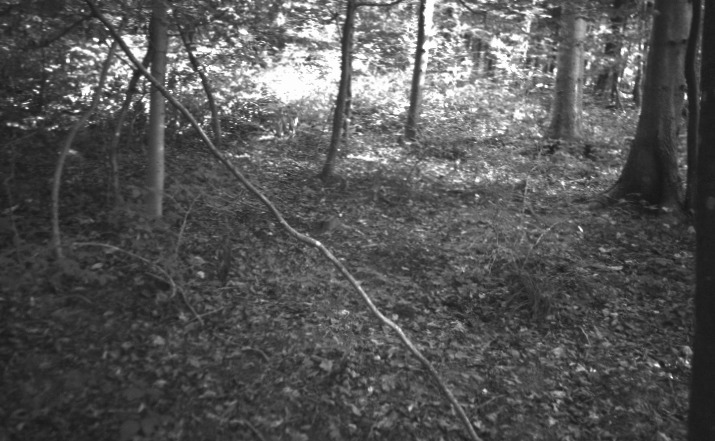} &
\begin{overpic}[width=0.16\textwidth]{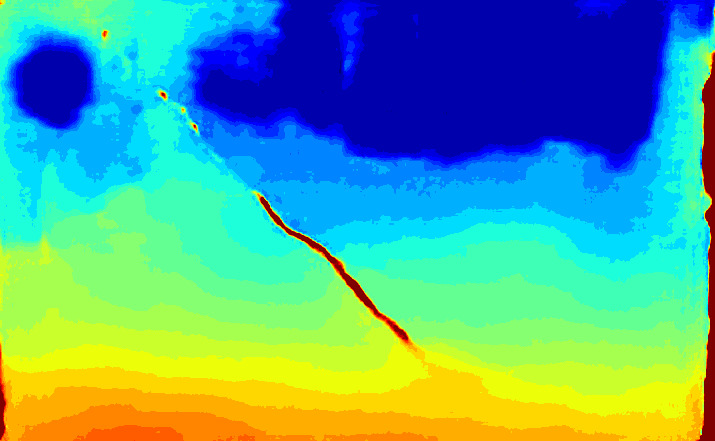} 
\put (2,48) {$\displaystyle\textcolor{white}{\textbf{Bad1: 33.32\%}}$}
\end{overpic} &
\begin{overpic}[width=0.16\textwidth]{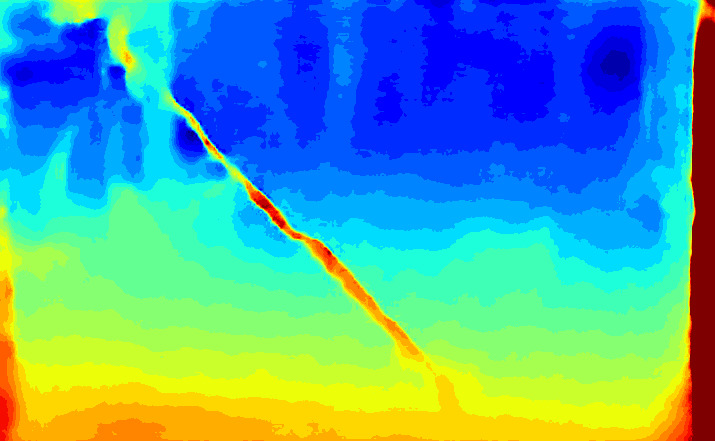} 
\put (2,48) {$\displaystyle\textcolor{white}{\textbf{Bad1: 16.31\%}}$}
\end{overpic} \\
\end{tabular}
\caption{Adaptation results for \dispnet{} on Middlebury v3 \cite{MIDDLEBURY_2014} (top) ETH3D dataset \cite{schoeps2017cvpr} (bottom). From left to right input (left) image, disparity maps predicted from network before any adaptation and after fine tuning with our adaptation technique. The bad1 error is overimposed on each map.}
\label{fig:qualitative_stereo}
\end{figure}

Finally we show some qualitative results, concerning both stereo and depth-from-mono networks, on the Middlebury v3 \cite{MIDDLEBURY_2014} and ETH3D \cite{schoeps2017cvpr} datasets. \autoref{fig:qualitative_mono} shows examples of depth maps obtained by \textit{monodepth} pre-trained on CityScapes  \cite{cordts2016cityscapes} before and after adaptation by our technique. The overall quality of the maps is greatly improved by the adaptation step, which is also vouched by the drastic drop of the absolute error reported in the Figure. We show similar results for \dispnet{} on \autoref{fig:qualitative_stereo}: the column labeled as \emph{No Adaptation} concerns predictions obtained by the model pre-trained on FlyingThings3D while the \emph{Adaptation} column deals with the results obtained after fine-tuning by our unsupervised adaptation approach. Results indicate clearly how our proposal can successfully correct the prediction range and drastically reduce the percentage of wrong pixels. 

Additional qualitative results are provided as supplementary material, in form of video sequences.

\section{Conclusion}
In this work, we have presented an effective methodology to fine-tune depth regressors based on CNNs towards brand-new environments by relying only on image pairs from the target domain. Through an extensive experimental evaluation, we have discussed the effectiveness of the different components of our method as well as proved its superior performance in comparison to popular alternatives dealing with both depth-from-stereo and depth-from-mono.

Our experiments suggest that combining naively noisy labels obtained from two very different stereo algorithms does not improve performance. Recent works like \cite{batsos2018cbmv}, however, have shown how combining different disparity estimations while taking into account the associated confidence maps can result in more reliable predictions. We plan to include in our framework a similar procedure in order to obtain more reliable disparity measurements from multiple, noisy stereo algorithms. Moreover, throughout this work, we have considered an offline adaptation phase aimed at ameliorating a successive online inference phase. Yet, one may conjecture a further extension of this concept whereby the two phases get fused together so as to adapt the depth prediction model online to ever-changing environments as soon as new images are gathered. By doing so, one may achieve better accuracy as well as realize a dynamic inference process capable of seamless adaption to unforeseen scenarios, like, \eg, bad weather conditions in autonomous driving, which, nowadays, are hardly dealt with by both hand-crafted and learning-based methods aimed at estimating depth from images.
{Along this path, we would also point out the potential for improving the accuracy of the confidence scores assigned to disparity labels in an online manner, \eg, by \emph{self-paced} learning techniques or estimating confidence scores by the disparity regressor itself like in \cite{Klodt_2018_ECCV}.}
Eventually, the ideas and experiments proposed in this paper concern \textit{adaptation} of a pre-trained CNN model to new settings. However, we believe that it would be worth investigating whether and how our unsupervised learning framework may be deployed to train a depth prediction model from scratch without supervision.

\ifCLASSOPTIONcompsoc
  \section*{Acknowledgments}
\else
  \section*{Acknowledgment}
\fi

We gratefully acknowledge the support of NVIDIA Corporation with the donation of the Titan Xp GPU used for this research.
Thanks to Filippo Aleotti for its help with experiments on monocular depth estimation.

\ifCLASSOPTIONcaptionsoff
  \newpage
\fi

\bibliographystyle{IEEEtran}
\bibliography{adaptation}
\vskip 3pt plus -1fil
\begin{IEEEbiography}
[{\includegraphics[width=1in,height=1.25in,clip,keepaspectratio]{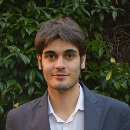}}]{Alessio Tonioni}
Received his PhD degree in Computer Science and Engineering from University of Bologna in 2019. 
Currently, he is a Post-doc researcher at Department of Computer Science and Engineering, University of Bologna.
His research interest concerns machine learning for depth estimation and object detection. 
\end{IEEEbiography}
\vskip 3pt plus -1fil
\begin{IEEEbiography}
[{\includegraphics[width=1in,height=1.25in,clip,keepaspectratio]{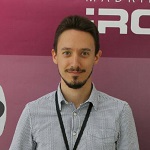}}]{Matteo Poggi}
received his PhD degree in Computer Science and Engineering from University
of Bologna in 2018. Currently, he is a Post-doc researcher at Department of Computer Science and Engineering, University of Bologna.
\end{IEEEbiography}
\vskip 3pt plus -1fil
\begin{IEEEbiography}
[{\includegraphics[width=1in,height=1.25in,clip,keepaspectratio]{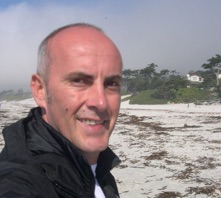}}]{Stefano Mattoccia}
received a Ph.D. degree in Computer Science Engineering from the University of Bologna in 2002. Currently
he is an associate professor at the Department
of Computer Science and Engineering of
the University of Bologna. His research interest
is mainly focused on computer vision, depth perception from images, deep learning and embedded computer vision. In these fields, he has authored about 100 scientific publications/patents. 
\end{IEEEbiography}
\vskip 3pt plus -1fil
\begin{IEEEbiography}
[{\includegraphics[width=1in,height=1.25in,clip,keepaspectratio]{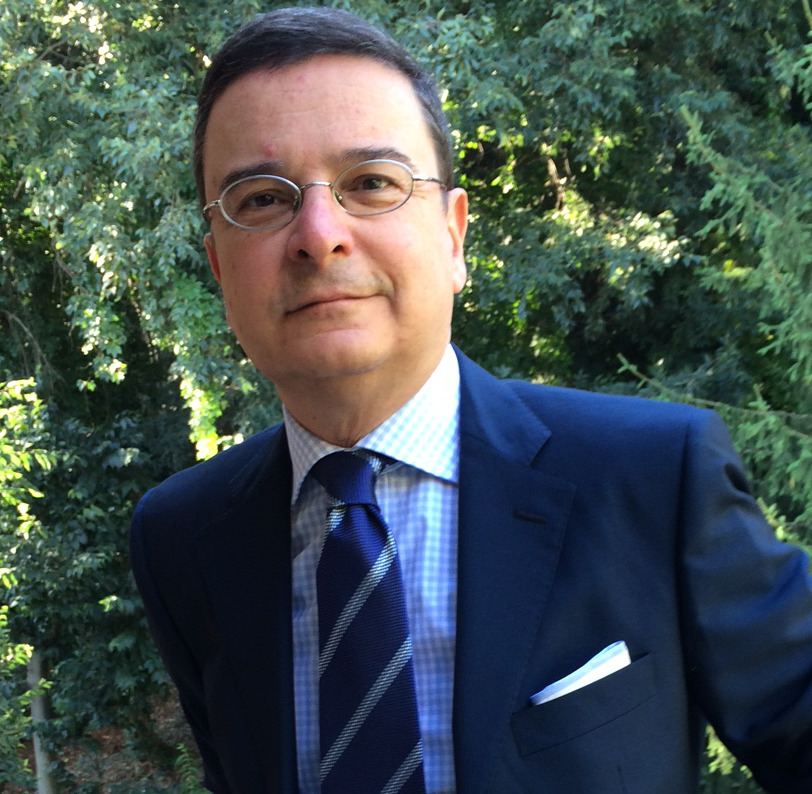}}]{Luigi Di Stefano}
received the PhD degree in electronic engineering and computer science from the University of Bologna in 1994. He is currently a full professor at the Department of Computer Science and Engineering, University of Bologna, where he founded and leads the Computer Vision Laboratory (CVLab). His research interests include image processing, computer vision and machine/deep learning. He is the author of more than 150
papers and several patents. He has been scientific consultant for major companies in the fields of computer vision and machine learning. 
He is a member of the IEEE Computer Society and the IAPR-IC.
\end{IEEEbiography}
\vskip 3pt plus -1fil

\end{document}